\documentclass[sigconf]{acmart}
\usepackage{enumitem}
\usepackage{booktabs}
\usepackage{svg}
\usepackage{subcaption}
\usepackage{amsmath}
\usepackage{multirow}
\usepackage{balance}
\DeclareMathOperator*{\argmaxA}{arg\,max}
\newlength\mylen
\AtBeginDocument{%
  }

\setcopyright{acmcopyright}

\copyrightyear{2023}
\acmYear{2023}
\setcopyright{rightsretained}
\acmConference[MM '23]{Proceedings of the 31st ACM International Conference on Multimedia}{October 29-November 3, 2023}{Ottawa, ON, Canada}
\acmBooktitle{Proceedings of the 31st ACM International Conference on Multimedia (MM '23), October 29-November 3, 2023, Ottawa, ON, Canada}

\makeatletter
\gdef\@copyrightpermission{
    \begin{minipage}{0.3\columnwidth}
    
    \href{https://creativecommons.org/licenses/by-nc-sa/4.0/}{\includegraphics[width=0.90\textwidth]{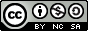}}

    \end{minipage}\hfill
    \begin{minipage}{0.7\columnwidth}
    
    \href{https://creativecommons.org/licenses/by-nc-sa/4.0/}{This work is licensed under the Creative Commons Attribution-NonCommercial-ShareAlike 4.0 International License.}
    
    \end{minipage}
    \vspace{5pt}
}
\makeatother

\begin{document}

\title[MM-AU]{MM-AU:Towards Multimodal Understanding of Advertisement Videos}

\author{Digbalay Bose}
\affiliation{%
  \institution{University of Southern California}
  \city{Los Angeles}
  \state{CA}
  \country{USA}
}
\email{dbose@usc.edu}

\author{Rajat Hebbar}
\affiliation{%
  \institution{University of Southern California}
  \city{Los Angeles}
  \state{CA}
  \country{USA}
}
\email{rajatheb@usc.edu}

\author{Tiantian Feng}
\affiliation{%
  \institution{University of Southern California}
  \city{Los Angeles}
  \state{CA}
  \country{USA}
}
\email{tiantianf@usc.edu}

\author{Krishna Somandepalli}
\authornote{The work was done while the author was at USC.}
\affiliation{%
 \institution{Google Research}
 \city{New York}
  \state{NY}
  \country{USA}}
\email{ksoman@google.com}

\author{Anfeng Xu}
\affiliation{%
 \institution{University of Southern California}
 \city{Los Angeles}
  \state{CA}
  \country{USA}}
\email{anfengxu@usc.edu}

\author{Shrikanth Narayanan}
\affiliation{%
  \institution{University of Southern California}
 \city{Los Angeles}
  \state{CA}
  \country{USA}}
\email{shri@ee.usc.edu}

\renewcommand{\shortauthors}{Digbalay Bose et al.}

\begin{abstract}
   Advertisement videos (ads) play an integral part in the domain of Internet e-commerce, as they amplify the reach of particular products to a broad audience or can serve as a medium to raise awareness about specific issues through concise narrative structures. The narrative structures of advertisements involve several elements like reasoning about the broad content (topic and the underlying message) and examining fine-grained details involving the transition of perceived tone due to the sequence of events and interaction among characters. In this work, to facilitate the understanding of advertisements along the three dimensions of topic categorization, perceived tone transition, and social message detection, we introduce a multimodal multilingual benchmark called \textbf{\textit{MM-AU}} comprised of 8.4 K videos (147hrs) curated from multiple web-based sources. We explore multiple zero-shot reasoning baselines through the application of large language models on the ads transcripts. Further, we demonstrate that leveraging signals from multiple modalities, including audio, video, and text, in multimodal transformer-based supervised models leads to improved performance compared to unimodal approaches. 
\end{abstract}


\begin{teaserfigure}
\begin{centering}
  \includegraphics[width=0.80\textwidth]{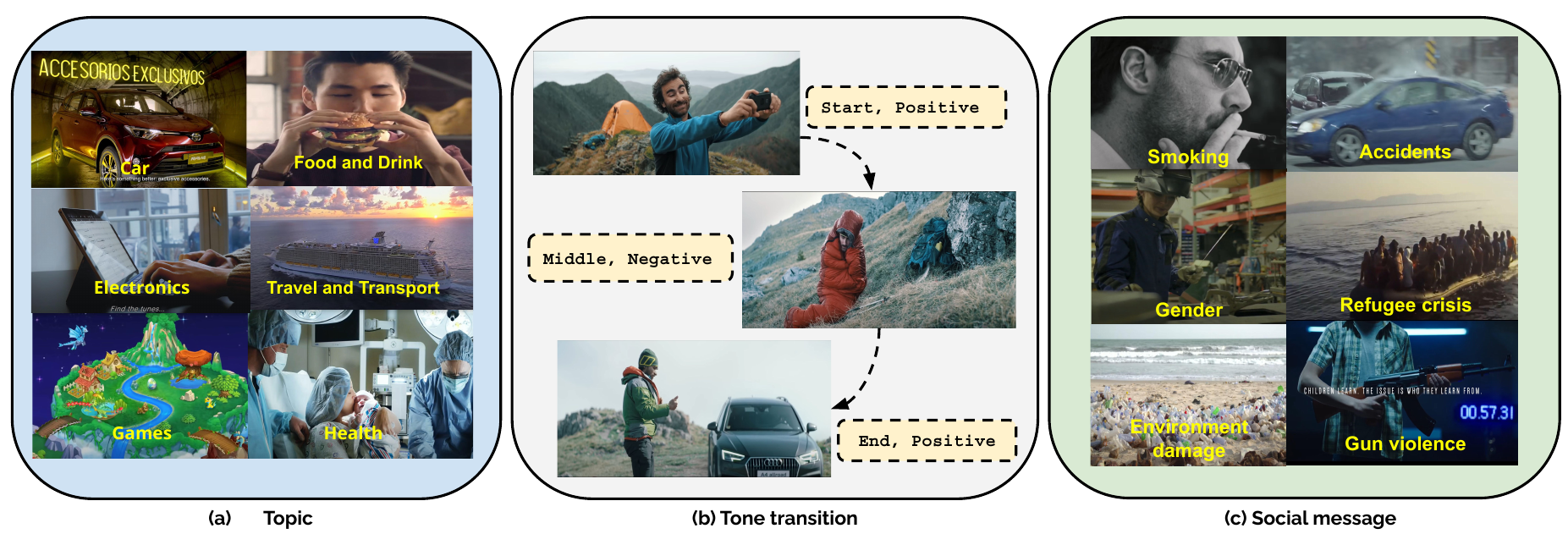}
  \caption{Schematic diagram showing illustrative examples of various tasks in the MM-AU (Multi-modal ads understanding) dataset. Multimodal understanding of ads along the lines of (a) Topic categorization (18 classes) (b) Tone transition (c) Social message detection, i.e. Absence/Presence of social message. Image sources: \cite{AOW}, \cite{cannes-lions}, \cite{Hussain2017AutomaticUO}}
  \label{introfig}
\end{centering}
\end{teaserfigure}


\maketitle

\section{Introduction}
Media content aims to portray rich human-centered narrative structures through various forms and outlets, including advertisements, movies, news, and increasingly as digital shorts on social media. 
As a primary media source for disseminating information, advertisements (ads for short) have been utilized to promote products or convey messages such as about extant social or political issues. The utility of advertisements as a media source has been amplified by the wide variety of platforms like radio, television, newspaper print, video streaming, and social networking sites, presenting significant influences, whether direct or indirect, to viewers from diverse backgrounds \cite{Pardun}. The rising importance of advertisements in the current socio-economic scenario is evident from the expected increase in media ad spending \cite{statisticsmovie} from 225.79 in 2020 to 322.11 billion dollars in 2024. 
\par 
Objectively understanding the rich content in ads, and their impact on the viewer experience and behavior is hence of great interest. However, in enabling computational media understanding \cite{CMI}, advertisements present unique challenges in the form of condensed narrative structures \cite{Kim2017WhyNA}. Due to their relatively short duration when compared to feature length movies, an advertisement video showcases a particular narrative structure in a tightly-integrated sequence with different formats, including slice-of-life \cite{Mick1987TowardAS}, drama \cite{Leong1994UsingDT}, and transformational \cite{Puto1984InformationalAT}. Further, reasoning about the narrative structure requires multi-scale understanding of the underlying topic and fine-grained elements, including the sequence of events 
(of characters and interactions) and related messages. As shown in Fig~\ref{introfig}, the key elements associated with the narrative structure of ads and related challenges are listed as follows:\\
\textbf{\underline{Topic}}: understanding enables personalized categorization and retrieval for customers along with key insights into the representation of genders \cite{google-diversity} and different demographic groups with respect to target classes like healthcare, retail, travel, etc. Topic categorization involves the handling of both inter and intra-topic diversity between the videos in terms of human-object interactions and a wide variety of items/products, as shown in Fig~\ref{introfig} (a). \\
\textbf{\underline{Tone transition}}: Affective tone associated with an advertisement video refers to the perceived feeling by a viewer \cite{Veer2008HowTT}. Associating the appropriate tone with an ad enhances its persuasiveness, thus enabling the associated brand to expand its reach to a wide range of customers. While the positive tone centers around optimistic elements associated with hope and success \cite{Brooks2020ExploringAO}, portrayals of negative tone are tied to sad narratives involving fear and suffering. However, due to the narrative structure, the perceived affective tone exhibits transitions during the duration of an advertisement video, accompanied by changes in visuals and background music.  In Fig~\ref{introfig} (b), the video starts on a positive note with a happy person taking a picture, followed by a perceived negative tone in the middle due to the suffering of the person. The advertisement ends on a positive note, with the person being saved by an incoming vehicle.\\
\textbf{\underline{Social message}}: Advertisements act as a major source of information about pressing social issues for consumers. Brands conveying messages about various social issues, including but not limited to gender inequalities, racial discrimination, and environmental conservation, are viewed favorably by consumers across different age groups \cite{Brooks2020ExploringAO}. In terms of advertisement videos, social message portrayal is characterized by a huge diversity in depiction, as shown in Fig~\ref{introfig} (c) due to underlying categories like smoking, accidents, gun violence etc.
\par
In this work, we introduce a multilingual multimodal benchmark called \textbf{\textit{MM-AU}} for the computational understanding of advertisement videos across the tasks of topic categorization, social message, and tone transition detection. Due to the inherent structure of the videos involving transitions in scenes, audio soundscapes, and diverse interactions among characters (text transcripts), we adopt a multi-modal approach to advertisement understanding. We outline our contributions below:
\begin{itemize}[leftmargin=*,labelsep=-\mylen]
    \item \textbf{\underline{Topic classification:}} We merge existing taxonomies for topic annotations from prior ads datasets and publicly available websites like Ads of the world \cite{AOW} to obtain a condensed set of topic labels for the advertisement videos.
    \item \textbf{\underline{Tone Transition detection:}} We introduce a novel benchmark task of tone transition detection in advertisement videos by obtaining crowdsourced feedback from human annotators.
    \item \textbf{\underline{Social message detection:}} We provide weak human expert-based labels for detecting the presence/absence of social messages in advertisement videos.
    \item \textbf{\underline{Language-based reasoning:}} We explore zero-shot baselines for the three benchmark tasks through applications of large-language models on ad transcripts.
    \item \textbf{\underline{Multimodal/Unimodal modeling:}} We provide multiple unimodal and multimodal baselines to benchmark the performance for the three tasks (topic classification, transition detection, and social message) and highlight future possibilities of explorations.
\end{itemize}
\section{Related work}
\textbf{Narrative understanding:} 
Narratives \cite{Fisher1987HumanCA} play an important role in enabling effective human communication and organizing the daily sequence of events. Advertisements centered on narratives \cite{Escalas1998ADVERTISINGNW} influence consumers by providing a concrete story arc centered around specific themes, protagonists and their actions. Kim et al. \cite{Kim2017WhyNA} introduced an integrated theory of understanding narratives in advertisements based on key variables like emotic response, ad hedonic value, ad credibility, and perceived goal facilitation. Lien et al. \cite{Lien2013NarrativeAT} explored narrative ads from the lens of persuasion and the relationship with different advertisement mediums - verbal or visual. In the realm of computational narrative understanding, prior works have focused on language-based approaches for marking high-level structures in short stories \cite{Li2017AnnotatingHS}, most reportable events (MRE) in Reddit comment threads \cite{Ouyang2015ModelingRE} and primary processes in movie scripts, newspaper articles, etc \cite{Boyd2020TheNA}.\\
\textbf{Affect modeling in videos:}
Advertisement brands tend to invoke emotional reactions \cite{Holbrook1984TheRO} in viewers by influencing their actions i.e., purchasing a particular product. In the domain of television commercials, a combination of physiological, symbolic, and self-report measures was explored in \cite{Micu2010MeasurableEH} to determine the emotional responses of viewers. The role of facial expressions in decoding the preferences of viewers, including purchase intent, and smile responses, has been explored through large-scale studies in \cite{McDuff2014PredictingAL}, \cite{Teixeira2014WhyWA}. Apart from facial expressions, the role of CNN-based audio-visual and EEG descriptors from the viewers has been explored in a multi-task setup \cite{Shukla2017AffectRI,Shukla2017EvaluatingCV,Shukla2019RecognitionOA} for arousal and valence prediction in advertisement videos. Existing video-based affect datasets like DEAP \cite{Koelstra2012DEAPAD}, VideoEmotion \cite{Jiang2014PredictingEI} also focused on single arousal, valence, and dominance ratings as well as discrete emotion labels for music and user-generated videos, respectively.\\
In the domain of continuous affect modeling, datasets with frame-level annotations have been introduced across a wide variety of domains, including naturalistic and induced clips (HUMAINE \cite{DouglasCowie2007TheHD}), movies (COGNIMUSE \cite{Zlatintsi2017COGNIMUSEAM}, LIRIS-ACCEDE \cite{Baveye2015LIRISACCEDEAV}) and online videos (EEV \cite{Sun2020EEVDP}). Further extensions of continuous affect modeling based on independent and self-reports have been explored for daily emotional narratives in SENDv1 dataset \cite{Ong2019ModelingEI}. For advertisements, climax annotations (presence + rough timestamps) were provided by human annotators on a subset of the Video Ads dataset \cite{Hussain2017AutomaticUO} in \cite{Ye2018StoryUI} along with climax-driven modeling strategies to predict sentiment labels at the video level. 
In our proposed benchmark \textbf{\textit{MM-AU}}, based on the standard definition in \cite{Brooks2020ExploringAO}, we ask human annotators to denote the perceived tone in the advertisement video across segments approximately marking the start, middle, and end. The perceived tone transition enables tracking of the narrative dynamics in advertisement videos by considering the interactions between different modalities, including audio, visual, and narrations/spoken interactions (through transcripts).\\
\textbf{Advertisement datasets:} 
\begin{table*}[h!]
\centering
\resizebox{0.99\textwidth}{!}{
\begin{tabular}{|c|c|c|c|c|c|c|c|c|}
\hline
\textbf{Dataset}       & \textbf{Annotation type} & \textbf{Duration} & \textbf{\#Samples} & \textbf{\#Shot} & \textbf{\#Class}                     & \textbf{Modalities}      & \textbf{Languages}                   & \textbf{Tasks}                      \\ \hline
Video Ads Dataset (I)  & H                        & NA                & 64832              & NA              & 38 (T), 30 (S), AR (OE), H(2), Ex(2) & Images                   & English                              & Image level classification          \\ \hline
Video Ads Dataset (V)  & H                        & 144.87            & 3477               & NA              & 38 (T), 30 (S), AR (OE), H(2), Ex(2) & Video                    & English                              & Video level classification          \\ \hline
Tencent-AVS            & H                        & 142.1h            & 12k                & 121.1k          & 25 (Pr), 34 (St), 23 (Pl)            & Video/Audio/ASR/OCR              & Chinese and English                  & Scene level classification          \\ \hline
Ads-persuasion dataset & H + AL                   & NA                & 3000               & NA              & 21 (PS)                              & Images                   & English                              & Image level classification          \\ \hline
E-MMAD                 & SG descriptions          & 1021.6 h          & 120984             & NA              & 4863 (PC)                            & Video                    & Chinese and English                  & Video level captioning              \\ \hline
\textbf{MM-AU}         & \textbf{H + SA}          & \textbf{147.8h}   & \textbf{8399}      & \textbf{216.4k} & \textbf{18 (T), 3 (Tone), 2 (SM)}    & \textbf{Video/Audio/ASR} & \textbf{Multilingual (65 languages)} & \textbf{Video level classification} \\ \hline
\end{tabular}
}
\vspace{5mm}
\caption{Comparison of \textbf{\textit{MM-AU}} with other available advertisement datasets across different modalities.  \underline{Annotation type:} \textbf{H}: Human annotation, \textbf{AL}: Active Learning, \textbf{SA}: Semi-automatic, \textbf{SG}: Store generated. \underline{Duration:} \textbf{NA}: Not applicable for images; Mentioned in hours(h). \underline{\#Samples:} Number of video clips or images. \underline{\#Shot:} \textbf{NA}: Not applicable for images; Number of shots detected from all the video samples. \underline{\#Class:} \textbf{T:} Topic, \textbf{S:} Sentiment, \textbf{OE:} Open-Ended, \textbf{H:} Humor, \textbf{Ex:} Exciting, \textbf{Pr:} Presentation, \textbf{St:} Style, \textbf{Pl:} Place, \textbf{PS:} Persuasion strategy, \textbf{PC:} Product categories, \textbf{SM:} Social messages}
\label{Overview}
\end{table*}
While there has been progress in terms of movie understanding due to the introduction of large-scale multimodal datasets like Condensed Movies \cite{bain2020condensed}, MovieNet \cite{huang2020movienet}, MAD \cite{Soldan_2022_CVPR}, Movie-cuts \cite{Pardo2021MovieCutsAN}, MovieCLIP \cite{Bose_2023_WACV}, AVE \cite{argaw2022anatomy} and SAM-S \cite{Hebbar2023ADF}, only few datasets have focused on large-scale understanding of advertisements across broad and fine-grained content. Hussain et al. \cite{Hussain2017AutomaticUO} introduced the benchmark Video-Ads dataset to faciliate understanding of images and videos along the lines of broad topics, induced sentiments and action/intent reasoning. The images in the Video-Ads dataset were utilized in \cite{Singla2022PersuasionSI} for computational modeling of persuasion across 21 categories in marketing domain. Regarding large scale ads understanding, Tencent-AVS dataset \cite{Jiang2022TencentAA} was proposed to enable multi-modal scene level categorization into semantic classes like presentation, places and styles. While the previously mentioned datasets focused on classification tasks, E-MMAD \cite{Zhang2022AttractMT} introduced the task of informative caption generation from advertisements across 120k e-commerce videos.
\par 
\textbf{\textit{MM-AU}}, our curated multilingual dataset utilizes videos from Ads-of-the-world \cite{AOW} along with a subset from Video-Ads dataset and an in-house video catalog from Cannes Lion archive \cite{cannes-lions}. We provide 18 broad topic categories by combining existing taxonomies (Cannes Lion, Ads-of-the-world and Video-Ads dataset). Further, we rely on human expert annotators to label transitions in perceived tone along with the absence/presence of social messages in 8.4K advertisement videos. A comparative overview of \textbf{\textit{MM-AU}} and other advertisement datasets is shown in Table \ref{Overview}.\\
\textbf{Semantic video understanding:}
Existing large-scale video datasets including Kinetics\cite{Carreira2017QuoVA}, Moments-in-time \cite{Monfort2018MomentsIT}, ActivityNet \cite{Heilbron2015ActivityNetAL}, AVA \cite{Gu2017AVAAV} have focused mainly on classifying entity driven actions from in-the-wild short videos. Higher-level semantic labels beyond actions like topics, concepts, events and video types were exploited for large-scale video-level categorization in datasets like Youtube-8M \cite{AbuElHaija2016YouTube8MAL}, Holistic-visual understanding (HVU) \cite{diba2020large} and 3MASSIV \cite{Gupta20223MASSIVMM}. Our proposed benchmark \textbf{\textit{MM-AU}} explores the domain of semantic video understanding in ads by considering broad categories like topic, presence/absence of social message, and fine-grained affective labels of perceived tone transition.\\
\textbf{Multimodal representation learning:} Multimodal representation learning \cite{Liang2022FoundationsAR} centers around the fusion of information from different modalities at multiple scales, including early, late, and mid-fusion. Prior works related to ads have utilized multimodal-LSTMs \cite{Vedula2017MultimodalCA}, segment-level autoencoders \cite{SomandepJointenc} or joint cross-modal embedding \cite{Ye2019InterpretingTR} approaches for learning multimodal representations for a variety of tasks. With the advent of transformer \cite{Transformers} based multimodal models like PerceiverIO \cite{Jaegle2021PerceiverIA}, attention bottlenecks\cite{nagrani2021attention}, and VATT \cite{Akbari2021VATTTF}, information fusion at the input token space, followed by joint encoders, have become more prevalent. A multi-task attention-based approach was explored in \cite{LookReadFeel} for jointly predicting topic and sentiment labels associated with advertisement images. A NextVLAD \cite{Lin2018NeXtVLADAE} based approach \cite{Weng2021AMF} combined with global-local attention was utilized for predicting scene-specific labels in the Tencent \cite{Jiang2022TencentAA} ads benchmark dataset.
\section{MM-AU dataset}
\subsection{Data sources}
\begin{figure}[h!]
    \centering
    \includegraphics[width=0.7\columnwidth]{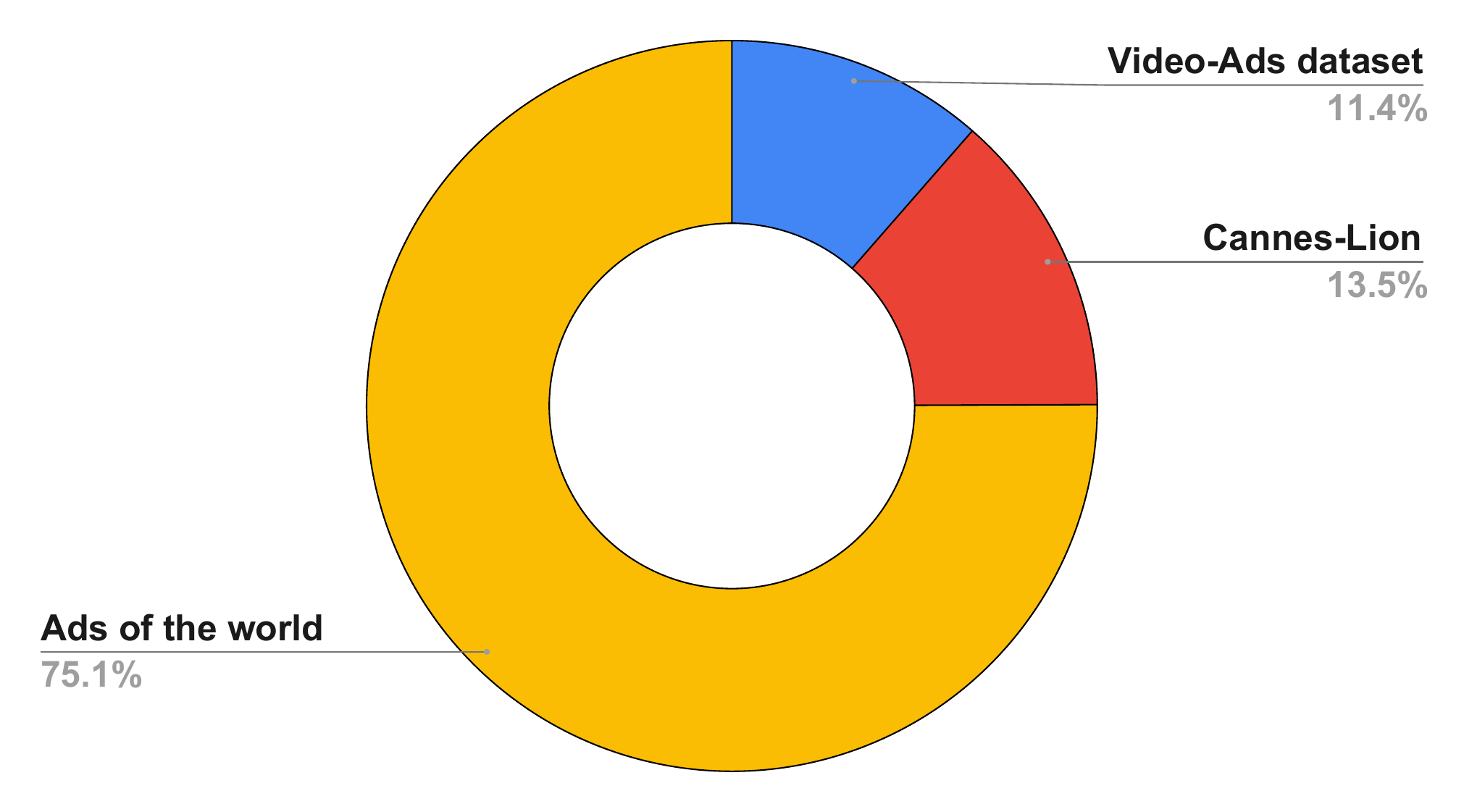}
    \caption{Share of different ad sources in MM-AU dataset. Ads of the world (6304 videos), Cannes Lion (1135), Video-Ads dataset (960) }
    \label{ads_sources}
\end{figure}

We consider multiple ads-specific sources for curating our proposed MM-AU dataset. As a primary source, we consider Ads-of-the-world (AOW) \cite{AOW} video hosting website since it contains a richly-curated catalog of ads in various formats like film, print, digital, and video spanning across multiple countries. As auxiliary sources, we consider additional videos from the Cannes Lion Film Festival \cite{cannes-lions} archive and Video-Ads dataset \cite{Hussain2017AutomaticUO}. We filter the videos based on unique video ids associated with their public links to ensure no duplicates across three sources. 
The share of different sources in curating the combined list of 8399 advertisement videos is shown in Fig~\ref{ads_sources}
We employ a semi-automatic process for tagging the advertisement videos with broad topic categories. For the detection tasks of tone transition and social message, we use Amazon Mechanical Turk \footnote{https://www.mturk.com/} to obtain responses from a pool of 36 human annotators. 
For selecting a pool of workers with the requisite expertise, we hosted an initial pilot study where the workers are instructed to mark the tone transition labels and presence/absence of social message in the given set of videos. Further, in the final annotation phase, three annotators independently annotate each sample for the tone-transition and social message detection tasks. An outline of the annotation process associated with tone transition and social message detection is provided in the Supplementary (Section 2.2). 
The annotation process details associated with the respective tasks are listed below:\\
\textbf{\underline{Tone transition:}} The annotators are instructed to mark the perceived tone labels associated with the start, middle, and ending segments of the advertisement videos. To reduce the burden associated with the task, no instructions are provided to mark the timestamps associated with the respective segments. Based on the tone definition considered in \cite{Brooks2020ExploringAO}, we provide the following descriptions to aid the annotation process:
\begin{itemize}[leftmargin=*,labelsep=-\mylen]
    \item \textbf{Positive tone:} An advertisement video segment has a positive tone if it contains: \textit{optimistic elements portraying hope and success} or \textit{positive imagery based on uplifting music and visuals}. Examples include girls overcoming negative stereotypes and succeeding in sports or a blind person being able to navigate easily through city roads by using an app.
    \item \textbf{Negative tone:} An advertisement video segment has a negative tone if it contains: \textit{sad narrative showing suffering, fear, destruction} or \textit{depressing music and distressing visuals}. Salient themes associated with negative tone include domestic violence, environmental damage, human trafficking, crisis from wars etc.
\end{itemize}
If a segment does not contain the above-mentioned characteristics, the annotators are instructed to mark the perceived tone as neutral. To determine the reasoning involved in marking the tone labels associated with the segments, the annotators are also asked to provide explanations regarding their choices. \\
\textbf{\underline{Social message detection:}} For social message detection, the annotators are instructed to check for the absence/presence of social messages in the given video. Based on the social message framing in ads \cite{Brooks2020ExploringAO}, we provide the following definition to guide the annotation process:
\begin{itemize}[leftmargin=*,labelsep=-\mylen]
    \item An advertisement video has a social message if it provides awareness about any social issue. Examples include \textit{gender equality}, \textit{drug abuse}, \textit{police brutality}, \textit{workplace harassment}, \textit{domestic violence}, \textit{child labor}, \textit{homelessness}, \textit{hate crimes} etc.
\end{itemize}
To simplify the annotation process, we ask the annotators to mark Yes/No for indicating the presence/absence of social messages in the videos instead of marking the exact categories in the curated list of social issues \cite{ciment2006social}.
\\
\textbf{\underline{Topic categorization:}}
We annotate topic categories using the existing taxonomies from Ads-of-the-world (AOW), Cannes Lions Film Festival \cite{cannes-lions}, and Video-Ads \cite{Hussain2017AutomaticUO} datasets. We denote the taxonomies associated with Cannes Lions Film Festival and Video-Ads datasets as Cannes-coding [\textcolor{purple}{\textbf{CC}}] and Video-Ads [\textcolor{blue}{\textbf{VA}}] coding schemes.
We extract the available tags associated with 6304 videos in Ads-of-the-world [\textcolor{red}{\textbf{AOW}}] and retain the top 40 tags based on frequency. Then we manually merge the filtered topic tags from AOW with similar labels in Cannes-coding  [\textcolor{purple}{\textbf{CC}}] and Video-Ads [\textcolor{blue}{\textbf{VA}}] coding schemes. Some examples of merged labels from different sources are listed as follows with the final parent topic category:
\begin{itemize}[leftmargin=*,labelsep=-\mylen]
    \item \textbf{Publications media:} Media \& Publications  [\textcolor{purple}{\textbf{CC}}]; Media and arts [\textcolor{blue}{\textbf{VA}}]; TV Promos, Music, Media, Movies [\textcolor{red}{\textbf{AOW}}]
    \item \textbf{Games:} Games and toys [\textcolor{blue}{\textbf{VA}}]; Gaming [\textcolor{red}{\textbf{AOW}}]
    \item \textbf{Sports:} Sports equipment and activities [\textcolor{blue}{\textbf{VA}}]; Sports [\textcolor{red}{\textbf{AOW}}]
    \item \textbf{Clothing:} Clothing, Footwear \& Accessories [\textcolor{purple}{\textbf{CC}}]; Clothing and accessories [\textcolor{blue}{\textbf{VA}}]; Personal Accessories [\textcolor{red}{\textbf{AOW}}]
\end{itemize}
A detailed list of mapping between the AOW, CC and VA coding schemes is included as part of the Supplementary (Section 2.1). Our final merged topic taxonomy consists of 18 categories as follows:
\begin{itemize}[leftmargin=*,labelsep=-\mylen]
    \item \textit{Games, Household, Services, Misc, Sports, Banking, Clothing, Industrial and agriculture, Leisure, Publications media, Health, Car, Electronics, Cosmetics, Food and Drink, Awareness, Travel and transport, Retail}
\end{itemize}
\textbf{\underline{Dataset Filtering:}} During the annotation process, we employ certain checks to maintain the quality of the annotated data. We reject those tone transition annotations with very short explanations (single words) or long generic descriptions of ads copied from the internet. Further, we also flag tone-transition annotations with the copied content across the start, middle, and end segments. For topic categorization, we merge categories with low frequencies, i.e., Alcohol and Restaurant, into the broad category of Food and Drink.
\begin{figure}[h!]
    \centering
    \includegraphics[width=0.6\columnwidth]{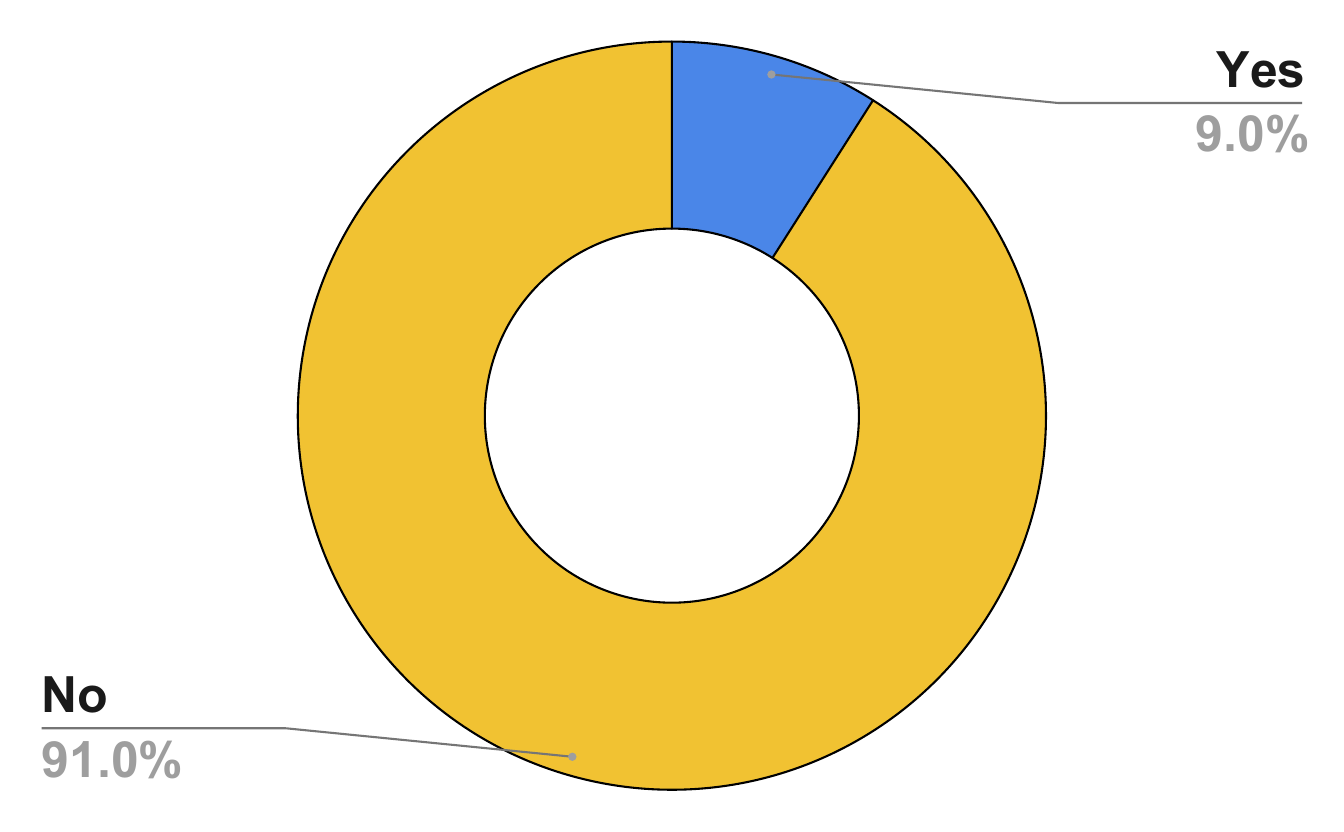}
    \caption{Distribution of the social message absence (No) and presence (Yes) labels in MM-AU}
    \label{Social_message_majority}
\end{figure}
\subsection{Dataset analysis}
\begin{figure*}[h!]
  \centering
  \includegraphics[width=0.8\textwidth]{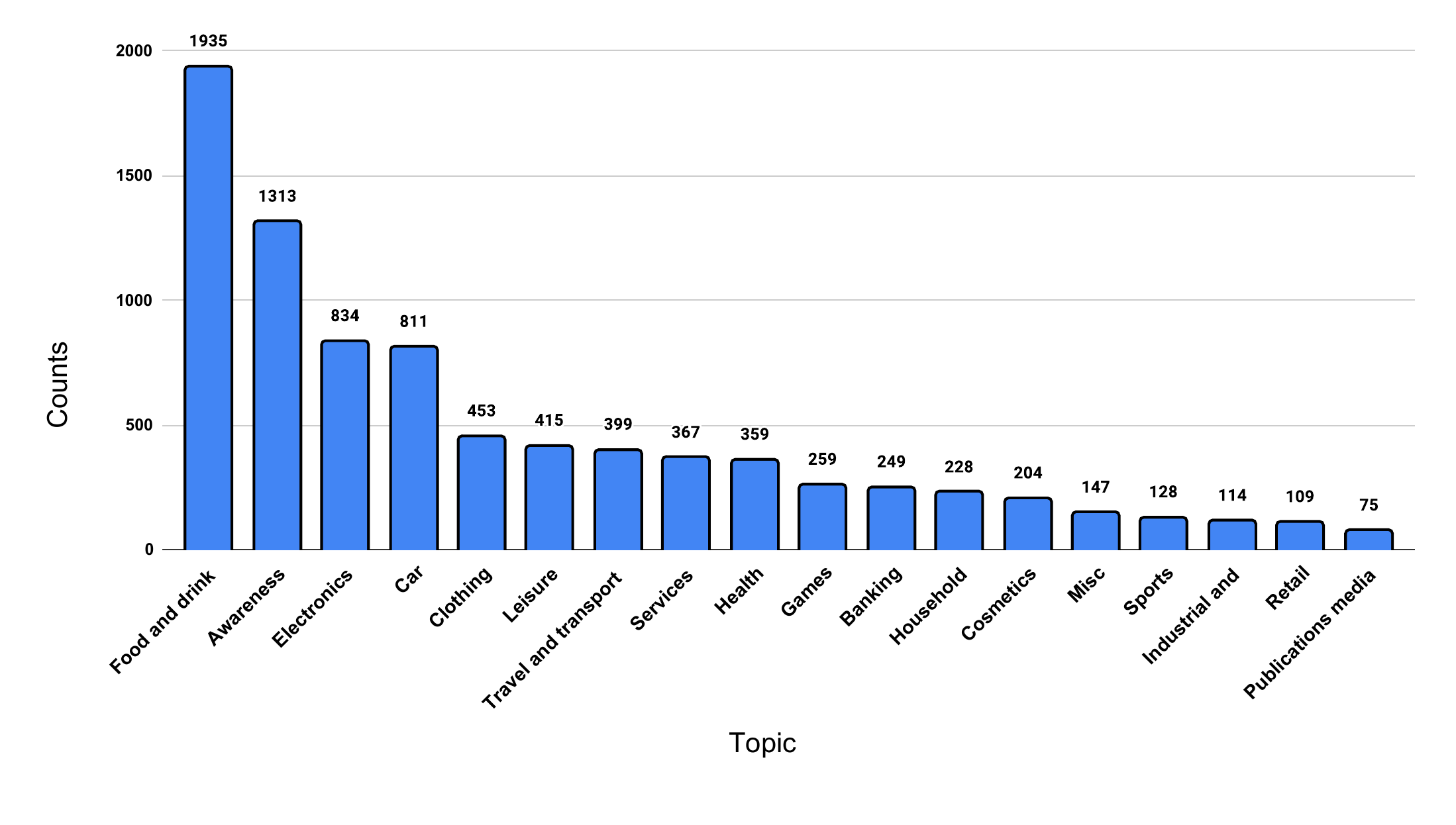}
  \caption{Distribution of topics in MM-AU dataset}
  \label{topics}
\end{figure*}
\begin{figure*}
\begin{subfigure}{.33\textwidth}
  \centering
  \includegraphics[width=.8\linewidth]{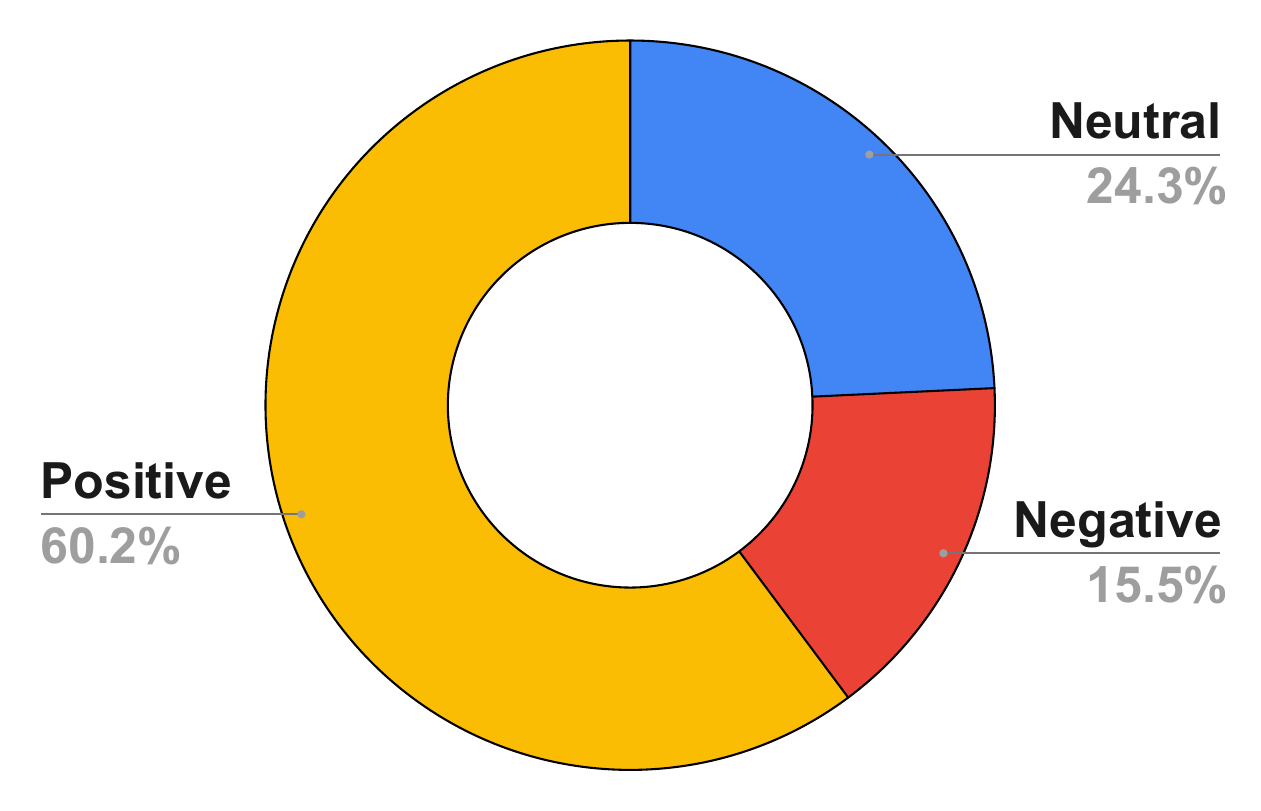}
  \caption{Start segment}
  \label{start_tone}
\end{subfigure}%
\begin{subfigure}{.33\textwidth}
  \centering
  \includegraphics[width=.8\linewidth]{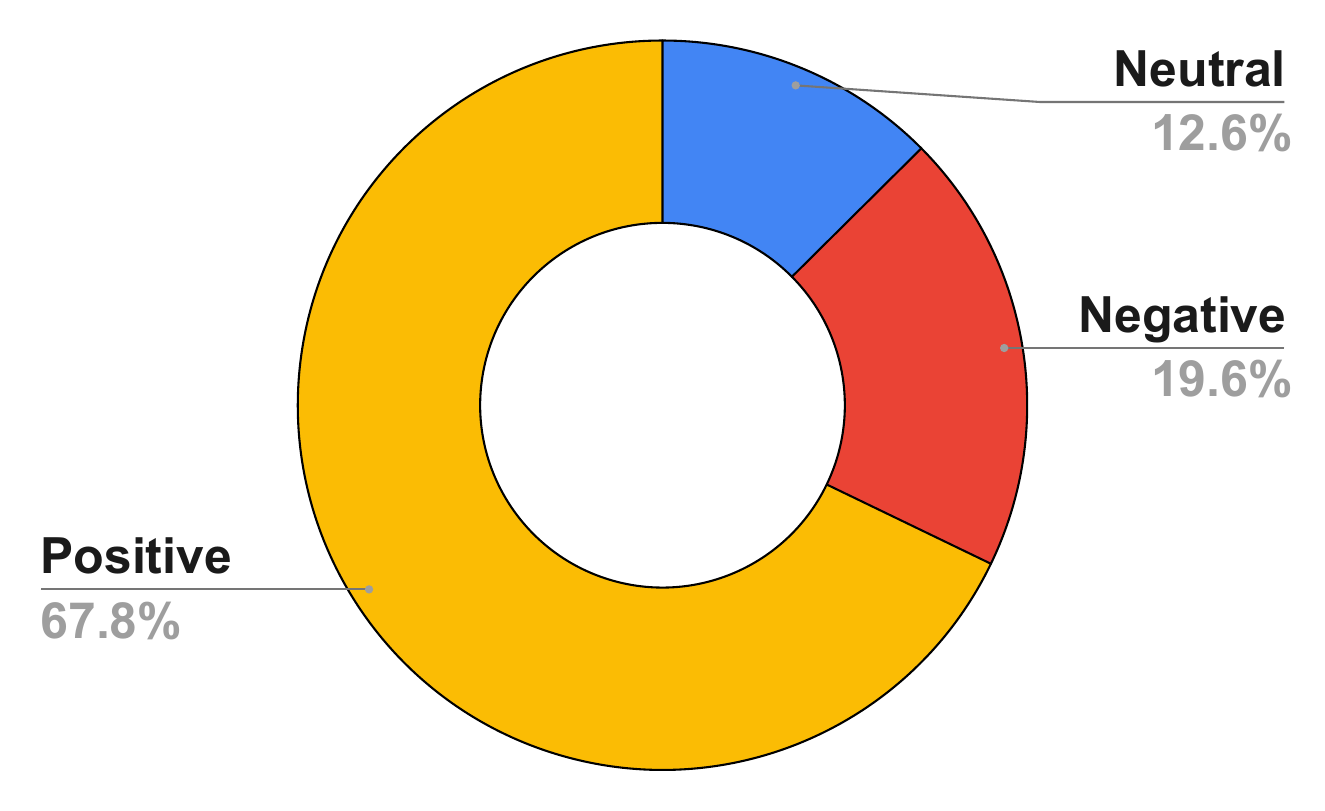}
  \caption{Middle segment}
  \label{mid_tone}
\end{subfigure}
\begin{subfigure}{.33\textwidth}
  \centering
  \includegraphics[width=.8\linewidth]{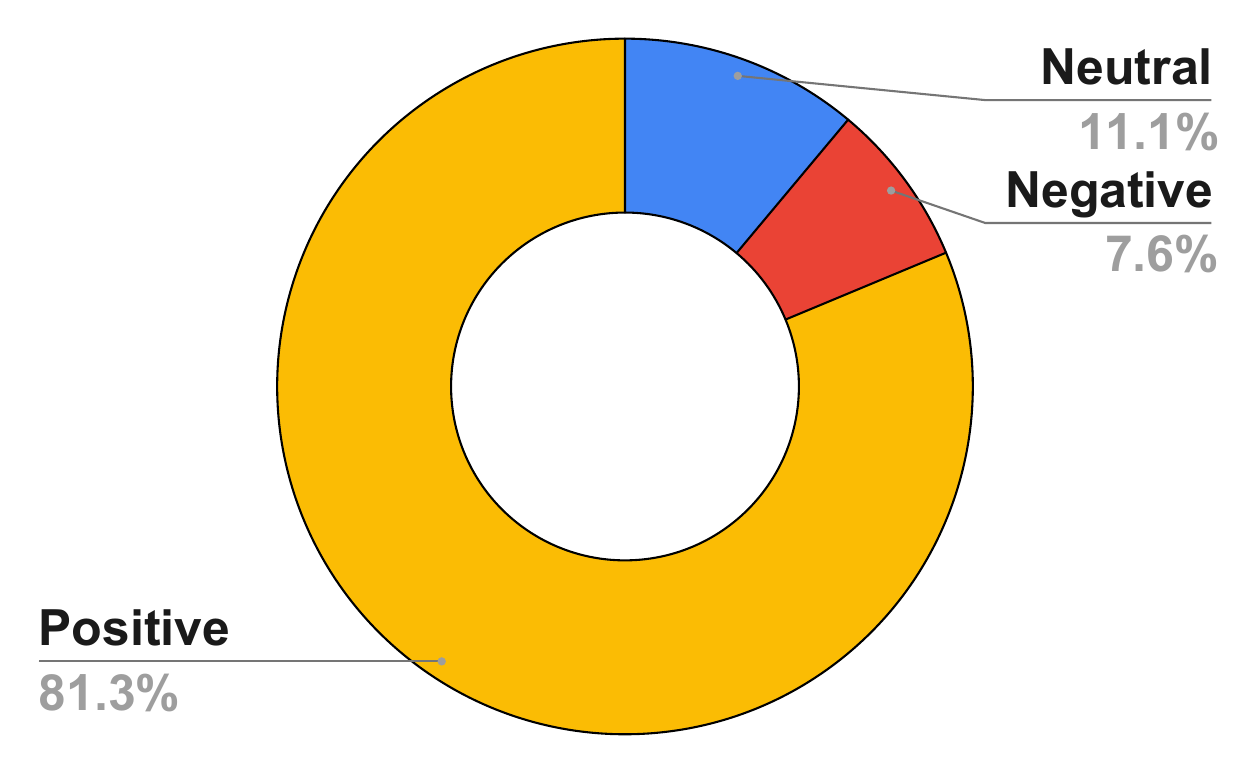}
  \caption{End segment}
  \label{end_tone}
\end{subfigure}
\caption{Distribution of majority perceived tone labels (among 3 annotators) across (a) start, (b) middle, and (c) ending segments in MM-AU dataset (8399 videos) }
\label{start_mid_end_tone}
\end{figure*}
MM-AU consists of 8399 annotated videos with a total of 147 hours of curated data. A detailed overview of MM-AU with total duration, number of tags, and year coverage is shown in Table \ref{Data_stats_table}.
\begin{table}[h!]
\begin{tabular}{@{}cc@{}}
\toprule
\textbf{Attribute} & \textbf{Value} \\ \midrule
\textbf{\#videos}           & 8399           \\
\textbf{\#explanations}     & 74970         \\
\textbf{\#topics}           & 18          \\
\textbf{\#social msg labels}       & 25197          \\
\textbf{\#tone labels}           & 75,591         \\
\textbf{\#duration}         & 147.79 hrs     \\
\textbf{\#avg duration}     & 63.35s         \\
\textbf{year}               & 2006-2019      \\
\textbf{\#annotators}       & 36             \\ 
\textbf{\#countries}       & 99             \\ \bottomrule
\end{tabular}
\caption{Data statistics of MM-AU dataset. \#social msg labels: total number of labels }
\label{Data_stats_table}
\end{table}
The distribution of topics is shown in Fig~\ref{topics}, with Food and Drink, Awareness, and Electronics being the top-3 dominant categories. In the case of perceived tone labels, we obtain a high majority agreement among annotators in marking the start (\textbf{91.2\%}), middle (\textbf{91.6\%}), and the ending (\textbf{94.5\%}) segments of the videos with perceived tone labels. Since annotating the presence/absence of social messages is a comparatively less subjective task than perceived tone labeling, we obtain a majority agreement (\textbf{99\%}) among the annotators. In terms of tone labels for start, middle, and end segments, we can see from Fig~\ref{start_mid_end_tone}, that the dominant perceived tone for the advertisements is positive, with its share rising from \textbf{60.2\%} (start) to \textbf{81.3\%} (end). This can be explained due to the fact that advertisements are primarily designed to persuade viewers to buy certain products or act toward certain social issues. However, from Fig~\ref{start_mid_end_tone}, we can see that the share of negative tone labels increases from \textbf{15.5\%} to \textbf{19.6\%} due to the narrative structure of the ads, where the middle segment portrays negative elements like human suffering, environmental damage, etc to set up the final conclusion. From Fig~\ref{Social_message_majority}, we can see that \textbf{9.0\%} of the videos, i.e. 759 contain social messages, as marked by \textbf{Yes} label. Out of 759 videos, \textbf{62.5\%} exhibit transition in perceived tone with the share of negative tone rising from \textbf{32.3\%} to \textbf{43.3\%} in the middle segments. Further intersectional views of social message presence with different topics and different segments (start, middle, and end) are included as part of the Supplementary (Figures 5 and 6). 
\section{Multimodal representative tasks}
For defining the multimodal representative tasks, we denote the audio, video, and text representations associated with the $ith$ video sample as $e_{a}$, $e_{v}$, $e_{t}$.\\
\textbf{\underline{Social message detection:}}
Based on the social message annotations (majority), the presence/absence of social message (SM) for $ith$ video is defined as:
\begin{equation}
  SM_{i} =
    \begin{cases}
      0 & \text{No (Absence of SM)}\\
      1 & \text{Yes (Presence of SM)}
    \end{cases}       
\end{equation}
Our aim is to learn a multimodal network $f_{SM}(e_{a},e_{v},e_{t})$ to predict social message presence/absence $\hat{SM}_{i}$ for the $ith$ video. The above definition results in 759 and 7640 videos marked with the presence  (1) and absence (0) of social messages, respectively.\\
\textbf{\underline{Tone transition:}}
Based on the start, middle, and end perceived tone labels (majority), we define the transition for $ith$ video as follows:
\begin{equation}
  Tr_{i} =
    \begin{cases}
      0 & \text{$Start_{i}=Mid_{i}=End_{i}$}\\
      1 & \text{else}
    \end{cases}       
\end{equation}
Our aim is to learn a multimodal network $f_{Tr}(e_{a},e_{v},e_{t})$ to predict binary tone transition $\hat{Tr}_{i}$ for the $ith$ video.
MM-AU dataset has 3854 and 4545 videos marked with Transition (1) and No Transition (0), respectively.\\
\textbf{\underline{Topic categorization:}}
For topic categorization, we aim to learn a multimodal network $f_{Topic}(e_{a},e_{v},e_{t})$ to predict $\hat{Topic}_{i}$ for the $ith$ video out of 18 target categories. 
\section{Proposed method}
\begin{figure*}[h!]
    \centering
    \includegraphics[width=0.7\textwidth]{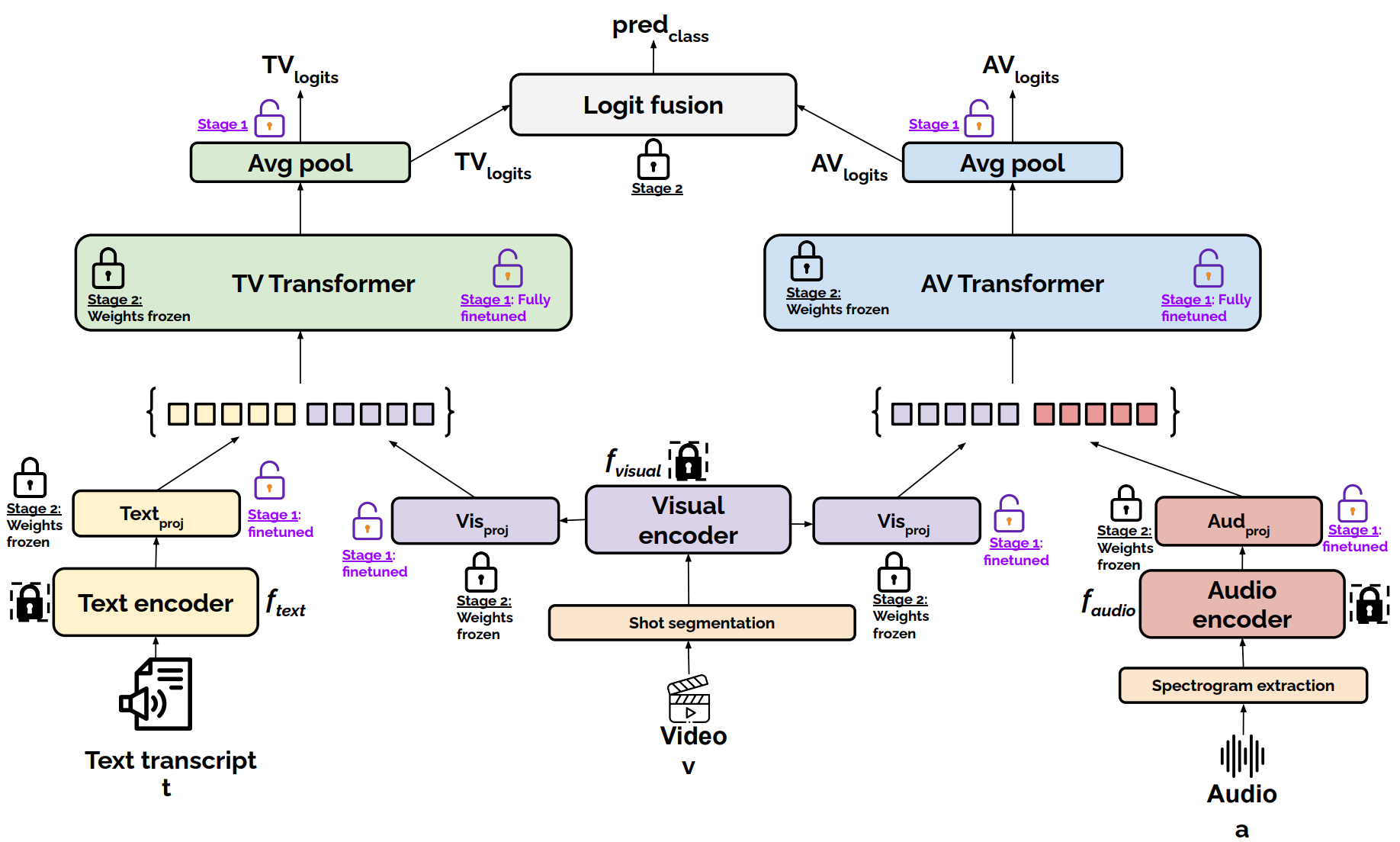}
    \caption{The proposed two-stage approach for fusing audio-visual (AV) and text-visual (TV) transformers. Stage 1: Full finetuning. Stage 2: Weights are completely frozen. The text, visual and audio encoders are kept completely frozen.}
    \label{Proposed_diagram}
\end{figure*}
We propose a two-stage method to combine multiple modalities i.e., audio, video, and text, in a transformer-based framework. For the transformer encoder, we use the PerceiverIO \cite{Jaegle2021PerceiverIA} architecture as part of our design choice due to its generalization capabilities to a wide variety of inputs. For $ith$ sample consisting of video ($\mathbf{v}$), audio ($\mathbf{a}$), text ($\mathbf{t}$),
the two-stage operation can be summarized as follows:\\
\textbf{\underline{Stage 1:}} Stage 1 involves complete finetuning of the T-V and A-V transformer encoder models indicated by $\mathbf{{Tx}_{TV}}$ and $\mathbf{{Tx}_{AV}}$. The text ($\mathbf{t}$), audio ($\mathbf{a}$), video 
 ($\mathbf{v}$) are passed through the respective encoders $f_{text}$, $f_{audio}$ and $f_{visual}$ respectively. The sequence of operations in Stage 1 can be summarized as follows:
 \begin{align}
    \begin{split}
     e_{text}&=Text_{proj}(f_{text}(t))\\
     e_{vis}&=Vis_{proj}(f_{visual}(v)) \\ 
     TV_{logits}&=\mathbf{AvgPool}(\mathbf{{Tx}_{TV}}([e_{text};e_{vis}])) \\
      e_{aud}&=Aud_{proj}(f_{audio}(a)) \\ 
      AV_{logits}&=\mathbf{AvgPool}(\mathbf{{Tx}_{AV}}([e_{vis};e_{aud}]))
    \end{split}
 \end{align}
 Here $Text_{proj}$, $Vis_{proj}$, $Aud_{proj}$ are linear projection layers used for mapping the outputs of respective modality-specific encoders to the same dimensions.[.;.] indicates concatenation along the temporal dimension.\\
\textbf{\underline{Stage 2:}}  In stage 2, we freeze the two transformer encoder models $\mathbf{{Tx}_{TV}}$ and $\mathbf{{Tx}_{AV}}$ and combine the respective logit outputs through the following strategies:
\begin{itemize}[leftmargin=*,labelsep=-\mylen]
    \item \textbf{A-max:} $pred_{class}=\argmaxA_i (TV_{logits}(i)+AV_{logits}(i))/2$
    \item \textbf{D-max:} $pred_{class}=\argmaxA_i max\{(TV_{logits}(i),AV_{logits}(i))\}$
\end{itemize}
Here $i \in \{1,...Class\}$, where $Class$ refers to the number of classes associated with the task. A detailed outline of our proposed approach is shown in Fig~\ref{Proposed_diagram}.
\section{Experiments}
\subsection{Experimental Setup}
For training, validation, and testing purposes, we consider a split of 5877 (70\%), 830 (10\%), and 1692 (20\%) videos. For the visual modality, we segment the videos into shots using \texttt{PySceneDetect}\footnote{https://github.com/Breakthrough/PySceneDetect} followed by extraction of frame-wise features at 4 fps using CLIP's \cite{Radford2021LearningTV} pretrained visual encoder ($f_{visual}$), \texttt{ViT-B/32}. Further, we average pool the frame-wise visual representations to obtain shot representations. \\
For the text modality, we use Whisper \cite{Radford2022RobustSR} multilingual model (\texttt{large}) to extract the transcripts. A detailed split of the languages available with the transcripts is shown in the Supplementary (Figure 9). We translate the multilingual transcripts to English using \texttt{GPT-4} \cite{OpenAI2023GPT4TR} (\textit{temp=0.02, max tokens=2048}) by providing the following translation-specific prompt: \textit{\textbf{Please provide an English translation of this transcript}}. We use the pretrained BERT\cite{Devlin2019BERTPO} model as the text encoder ($f_{text}$). For the audio modality, we use the Audio-Spectrogram transformer (AST) \cite{gong21b_interspeech} model ($f_{audio}$) pretrained on AudioSet \cite{Gemmeke2017AudioSA} for extracting features at 10-sec intervals with a step size of 512. We conduct our experiments in a distributed manner using the Pytorch \cite{Paszke2019PyTorchAI} framework on 4 2080ti GPUs. For evaluation metrics, we use accuracy and macro-F1.
\subsection{Language-based reasoning}
We investigate the zero-shot capabilities of foundational large language models \cite{Zhao2023ASO} by applying \texttt{GPT-4} \cite{OpenAI2023GPT4TR}, \texttt{Opt-IML} \cite{Iyer2022OPTIMLSL}, \texttt{Flan-T5} (XXL,XL,L) \cite{Chung2022ScalingIL} and \texttt{Alpaca} \cite{alpaca} on the translated transcripts. For zero-shot evaluation, we report the results on 1670 non-empty transcripts out of the test split of 1692 samples. For \texttt{GPT-4} we use the following task-specific prompts:
\begin{itemize}[leftmargin=*,labelsep=-\mylen]
    \item \textbf{SM:} \textit{An advertisement video has a social message if it provides awareness about any social issue. Example of social issues: gender equality, drug abuse, police brutality, workplace harassment, domestic violence, child labor, environmental damage, homelessness, hate crimes, racial inequality etc. Based on the given text transcript, determine if the advertisement has any social message. Please provide answers in \textbf{Yes} and \textbf{No.}}
    \item \textbf{TT:} \textit{Based on the given text transcript from the advertisement, determine if the advertisement has any transitions in tones. Possible tone labels are: positive, negative, and neutral. Please respond by saying \textbf{Transition} or \textbf{No transition}}
    \item \textbf{Topic:} \textit{Associate a single topic label with the transcript from the given set: <Topic list>}
\end{itemize}
Here SM, TT, and Topic refer to the benchmark tasks of Social message detection, tone transition, and topic categorization. <Topic list> refers to the condensed list of 18 topic categories curated for MM-AU dataset. Further details about the prompting strategies are included as part of the Supplementary (Section 3.1).
\subsection{Unimodal and Multimodal baselines}
For the supervised unimodal baselines we consider the following model choices:
\begin{itemize}[leftmargin=*,labelsep=-\mylen]
    \item \textbf{LSTM \cite{Hochreiter1997LongSM}:} 2 layers and hidden dimension = 256
    \item \textbf{MHA \cite{Transformers}:} 4 layers, 4 heads and hidden dimension = 256
\end{itemize}
For multimodal models, we use Perceiver IO structure \cite{Jaegle2021PerceiverIA} as the attention-based transformer encoder to explore combinations of different paired modalities i.e., audio-visual ($\mathbf{Tx_{AV}}$), text-visual ($\mathbf{Tx_{TV}}$), audio-text ($\mathbf{Tx_{AT}}$). For the perceiver IO blocks, we adopt a lightweight structure composed of 4 encoder layers, 16 latent vectors, 8 heads, and a hidden latent dimensionality of 256. We use binary cross-entropy for social message and tone transition detection tasks and multi-class cross-entropy for topic categorization. For training the unimodal and multimodal models, we use a batch size of 16 with Adam \cite{kingma2014adam} or AdamW \cite{Loshchilov2017DecoupledWD} as optimizers and $lr \in \{1e-4, 1e-5\}$. While training the supervised models, we fix the maximum sequence lengths for visual(shots), audio, and text modalities at 35, 14, and $\{256,512\}$, respectively. During multimodal fusion, when text is missing in the transcripts due to the absence of speech, we replace the text with a string of \texttt{[MASK]} tokens. A detailed breakdown of model-wise hyperparameters is included in the Supplementary (Section 3.2). 
\begin{table}[h!]
\resizebox{\columnwidth}{!}{
\begin{tabular}{|cc|cc|cc|cc|}
\hline
\multicolumn{2}{|c|}{\textbf{Configurations}}                & \multicolumn{2}{c|}{\textbf{SM}}                & \multicolumn{2}{c|}{\textbf{TT}}                & \multicolumn{2}{c|}{\textbf{Topic}}             \\ \hline
\multicolumn{1}{|c|}{\textbf{Model}}       & \textbf{Params} & \multicolumn{1}{c|}{\textbf{Acc}} & \textbf{F1} & \multicolumn{1}{c|}{\textbf{Acc}} & \textbf{F1} & \multicolumn{1}{c|}{\textbf{Acc}} & \textbf{F1} \\ \hline
\multicolumn{1}{|c|}{\textbf{GPT-4 \textcolor{red}{\cite{OpenAI2023GPT4TR}}}}       & $NA^{*}$             & \multicolumn{1}{c|}{\textbf{87.6}}         & \textbf{65.66}    & \multicolumn{1}{c|}{\textbf{58.56}}        & \textbf{58.33}       & \multicolumn{1}{c|}{\textbf{33.29}}       & \textbf{29.21}       \\ \hline
\multicolumn{1}{|c|}{\textbf{Flan-T5-XXL \textcolor{red}{\cite{Chung2022ScalingIL}}}} & 11B             & \multicolumn{1}{c|}{85.69}        & 62.7        & \multicolumn{1}{c|}{54.79}        & 44.15       & \multicolumn{1}{c|}{30.54}        & 24.23       \\ \hline
\multicolumn{1}{|c|}{\textbf{Flan-T5-XL \textcolor{red}{\cite{Chung2022ScalingIL}}}}  & 3B              & \multicolumn{1}{c|}{65.51}        & 49.31       & \multicolumn{1}{c|}{54.67}        & 42.39       & \multicolumn{1}{c|}{27.18}        & 24.1        \\ \hline
\multicolumn{1}{|c|}{\textbf{Alpaca  \textcolor{red}{\cite{alpaca}}}}      & 7B              & \multicolumn{1}{c|}{10.77}        & 10.56       & \multicolumn{1}{c|}{46.88}        & 39.1        & \multicolumn{1}{c|}{11.19}        & 11.68       \\ \hline
\multicolumn{1}{|c|}{\textbf{Opt-IML \textcolor{red}{\cite{Iyer2022OPTIMLSL}}}}     & 1.3B            & \multicolumn{1}{c|}{37.07}        & 32.49       & \multicolumn{1}{c|}{54.37}        & 35.22       & \multicolumn{1}{c|}{22.22}        & 19.08       \\ \hline
\multicolumn{1}{|c|}{\textbf{Flan-T5-L \textcolor{red}{\cite{Chung2022ScalingIL}}}}   & 780M            & \multicolumn{1}{c|}{8.32}         & 7.76        & \multicolumn{1}{c|}{54.43}        & 35.25       & \multicolumn{1}{c|}{26.82}        & 19.42       \\ \hline
\multicolumn{1}{|c|}{Random Baseline}      & \textbf{-}           & \multicolumn{1}{c|}{49.57}        & 39.61       & \multicolumn{1}{c|}{49.95}        & 49.88        & \multicolumn{1}{c|}{5.71}        & 4.71 \\ \hline
\multicolumn{1}{|c|}{Majority Baseline}      & \textbf{-}           & \multicolumn{1}{c|}{90.96}        & 47.63       & \multicolumn{1}{c|}{54.11}        & 35.11        & \multicolumn{1}{c|}{23.04}        & 2.08 \\ \hline
\end{tabular}
}
\caption{Zero shot performance comparison between various LLMs on MM-AU dataset.
Tasks: \textit{SM}: Social message detection, \textit{TT}: Tone transition, \textit{Topic}: Topic categorization. $NA^{*}:$  Information not available.  F1: Macro-F1. Best performing results are marked in bold for respective tasks.}
\label{llm}
\end{table} 
\section{Results}
\begin{table*}[h!]
\resizebox{0.9\textwidth}{!}{
\begin{tabular}{|ccccccccc|}
\hline
\multicolumn{3}{|c|}{\textbf{Configurations}}                                                                               & \multicolumn{1}{c|}{\textbf{Acc}} & \multicolumn{1}{c|}{\textbf{F1}} & \multicolumn{1}{c|}{\textbf{Acc}} & \multicolumn{1}{c|}{\textbf{F1}} & \multicolumn{1}{c|}{\textbf{Acc}}   & \textbf{F1}  \\ \hline
\multicolumn{1}{|c|}{\textbf{Model}}      & \multicolumn{1}{c|}{\textbf{Features}} & \multicolumn{1}{c|}{\textbf{Modality}} & \multicolumn{2}{c|}{\textbf{Social message}}                         & \multicolumn{2}{c|}{\textbf{Tone transition}}                        & \multicolumn{2}{c|}{\textbf{Topic categorization}} \\ \hline
\multicolumn{1}{|c|}{\textbf{Random}}     & \multicolumn{1}{c|}{NA}                & \multicolumn{1}{c|}{NA}                & \multicolumn{1}{c|}{49.57±0.28}   & \multicolumn{1}{c|}{39.61±0.28}  & \multicolumn{1}{c|}{49.95±0.30}   & \multicolumn{1}{c|}{49.88±0.30}  & \multicolumn{1}{c|}{5.71±0.16}      & 4.71±0.24    \\ \hline
\multicolumn{1}{|c|}{\textbf{Majority}}   & \multicolumn{1}{c|}{NA}                & \multicolumn{1}{c|}{NA}                & \multicolumn{1}{c|}{90.96}        & \multicolumn{1}{c|}{47.63}       & \multicolumn{1}{c|}{54.11}        & \multicolumn{1}{c|}{35.11}       & \multicolumn{1}{c|}{23.04}          & 2.08         \\ \hline
\multicolumn{9}{|c|}{\textbf{Unimodal}}                                                                                                                                                                                                                                                                                        \\ \hline
\multicolumn{1}{|c|}{\textbf{LSTM}}       & \multicolumn{1}{c|}{CLIP-S}            & \multicolumn{1}{c|}{V}                 & \multicolumn{1}{c|}{90.57±0.47}   & \multicolumn{1}{c|}{68.65±1.70}  & \multicolumn{1}{c|}{61.93±0.37}   & \multicolumn{1}{c|}{61.65±0.37}  & \multicolumn{1}{c|}{52.86±0.43}     & 36.48±1.58   \\ \hline
\multicolumn{1}{|c|}{\textbf{MHA}}        & \multicolumn{1}{c|}{AST}               & \multicolumn{1}{c|}{A}                 & \multicolumn{1}{c|}{89.07±2.02}   & \multicolumn{1}{c|}{55.33±3.96}  & \multicolumn{1}{c|}{59.78±1.56}   & \multicolumn{1}{c|}{58.60±0.96}  & \multicolumn{1}{c|}{25.11±1.39}     & 15.48±1.57   \\ \hline
\multicolumn{1}{|c|}{\textbf{MHA}}        & \multicolumn{1}{c|}{CLIP-S}            & \multicolumn{1}{c|}{V}                 & \multicolumn{1}{c|}{90.41±2.24}   & \multicolumn{1}{c|}{72.28±1.66}  & \multicolumn{1}{c|}{61.74±1.07}   & \multicolumn{1}{c|}{61.48±1.11}  & \multicolumn{1}{c|}{61.30±0.89}     & 47.74±1.27   \\ \hline
\multicolumn{9}{|c|}{\textbf{Multimodal}}                                                                                                                                                                                                                                                                                      \\ \hline
\multicolumn{1}{|c|}{$\mathbf{Tx_{AT}}$}      & \multicolumn{1}{c|}{AST + BERT}        & \multicolumn{1}{c|}{A + T}             & \multicolumn{1}{c|}{90.24±0.81}   & \multicolumn{1}{c|}{64.02±0.81}  & \multicolumn{1}{c|}{62.98±0.56}   & \multicolumn{1}{c|}{62.29±0.83}  & \multicolumn{1}{c|}{42.99±0.65}     & 30.77±0.96   \\ \hline
\multicolumn{1}{|c|}{\textbf{(1) $\mathbf{Tx_{AV}}$}}  & \multicolumn{1}{c|}{CLIP-S +AST}       & \multicolumn{1}{c|}{A + V}             & \multicolumn{1}{c|}{91.62±0.58}   & \multicolumn{1}{c|}{70.05±0.67}  & \multicolumn{1}{c|}{64.01±0.54}   & \multicolumn{1}{c|}{63.72±0.66}  & \multicolumn{1}{c|}{61.62±0.46}     & 48.67±0.64   \\ \hline
\multicolumn{1}{|c|}{\textbf{(2) $\mathbf{Tx_{TV}}$}}  & \multicolumn{1}{c|}{CLIP-S +BERT}      & \multicolumn{1}{c|}{T + V}             & \multicolumn{1}{c|}{\textbf{92.23±0.57}}  & \multicolumn{1}{c|}{\textbf{74.03±1.00}}  & \multicolumn{1}{c|}{63.96±0.99}   & \multicolumn{1}{c|}{63.48±0.84}  & \multicolumn{1}{c|}{63.27±0.59}     & 50.58±1.32   \\ \hline
\multicolumn{1}{|c|}{\textbf{A-Max(1,2)}} & \multicolumn{1}{c|}{CLIP-S +BERT +AST} & \multicolumn{1}{c|}{A + V + T}          & \multicolumn{1}{c|}{92.51±0.46}   & \multicolumn{1}{c|}{73.17±1.00}  & \multicolumn{1}{c|}{\textbf{65.05±0.36}}   & \multicolumn{1}{c|}{\textbf{64.67±0.33}} & \multicolumn{1}{c|}{\textbf{65.92±0.54}}     & \textbf{54.22±1.14}  \\ \hline
\multicolumn{1}{|c|}{\textbf{D-Max(1,2)}} & \multicolumn{1}{c|}{CLIP-S +BERT +AST} & \multicolumn{1}{c|}{A + V + T}          & \multicolumn{1}{c|}{92.52±0.46}   & \multicolumn{1}{c|}{73.21±0.98}  & \multicolumn{1}{c|}{65.01±0.39}   & \multicolumn{1}{c|}{64.63±0.32}  & \multicolumn{1}{c|}{65.51±0.58}     & 53.67±1.24   \\ \hline
\end{tabular}
}
\caption{Comparative results between different unimodal and multimodal models across different tasks: Social message and Tone transition, Topic categorization. \textbf{\textit{\underline{CLIP-S}}}: Shot level features extracted using CLIP. \textbf{\textit{\underline{Modality:}}} A: Audio, V: Visual, T: Text. Results are reported as an average of 5 runs with randomly selected seeds. Best performing results are marked in bold for respective tasks.}  
\label{exptable}
\end{table*}
\subsection{Language-based reasoning}
Based on results in Table \ref{llm}, we can see that \texttt{GPT-4} exhibits superior zero-shot performance (F1 and Accuracy) across all the tasks when compared to other large language models. Further, there is a trend towards improved model performance with model scaling, except for \texttt{Alpaca}. The poor performance of \texttt{Alpaca} can be attributed to a lack of self-instruct data associated with complex reasoning tasks from transcripts. Instruction finetuning coupled with model scaling improves the zero-shot performance (F1) of T-5 models from \textbf{49.31\%} to \textbf{62.7\%} and \textbf{42.39\%} to \textbf{44.15\%} for social message and tone transition tasks respectively.
\subsection{Unimodal and Multimodal baselines}
From Table \ref{exptable}, we can see that supervised unimodal models (MHA, LSTM) show improved performance as compared to simple random and majority baselines. In terms of unimodal models, MHA model trained on shot-level visual features (denoted by CLIP-S) perform far better than audio features (AST) in social message detection (\textbf{F1: 72.28} vs \textbf{F1: 55.33}) and topic categorization tasks.
(\textbf{F1: 47.74} vs \textbf{F1: 15.48}). For the tone transition task, MHA model trained on audio features (AST) shows close performance (\textbf{F1:58.60} vs \textbf{F1: 61.48}) as compared to visual features, due to the dependence of the tone transition task on the ambient music.  \\
For multimodal models, we observe that the fusion of text and visual modalities through a Perceiver-IO-based encoder ($\mathbf{Tx_{TV}}$) performs better (\textbf{F1:74.03}) for social message detection compared to audio-visual or audio-text fusion. This can be attributed to the presence of socially-relevant descriptors in the transcripts and video shots. However, the fusion of audio with visual signals ($\mathbf{Tx_{AV}}$) improves the performance in the tone-transition detection task (\textbf{F1: 63.72}). For topic categorization, we find that the fusion of text and visual modalities ($\mathbf{Tx_{TV}}$) performs better than other paired modalities (\textbf{F1:50.58}) due to topic-specific identifiers in transcripts and shots. 
\par
 Our proposed approach based on logit fusion strategies: Average-Max (\textbf{A-Max}) and Dual-Max (\textbf{D-Max}) exhibits similar performance across all tasks. We obtain gain for tone-transition (\textbf{F1:64.67}) based on the \textbf{A-Max} fusion strategy of $\mathbf{Tx_{AV}}$ and $\mathbf{Tx_{TV}}$ models.  In terms of class-wise metrics, \textbf{A-Max} fusion improves the transition class average F1-score to \textbf{61.28\%} as compared to $\mathbf{Tx_{AV}}$ (\textbf{60.72\%}) and $\mathbf{Tx_{TV}}$ (\textbf{60.18\%}). In the case of topic categorization, logit fusion through \textbf{A-Max} of $\mathbf{Tx_{AV}}$ and $\mathbf{Tx_{TV}}$ models results in the best performance (\textbf{F1: 54.22}). Further, \textbf{A-Max} fusion results in improvements over $\mathbf{Tx_{AV}}$ and $\mathbf{Tx_{TV}}$ across 15 topic categories (out of 18), with noticeable gains in the minority categories i.e., Retail ($\sim${\textbf{6.2\%}}), Industrial \& Agriculture ($\sim$\textbf{5\%}), Household ($\sim$\textbf{7\%}). Similar trends can be observed for \textbf{D-Max} fusion, with improvements obtained across 14 topic categories (out of 18). A detailed class-wise breakdown for topic categorization is included as part of the Supplementary (Figure 13).
\section{Conclusions}
 We introduced a multimodal multilingual ads dataset called MM-AU, and presented the core benchmark tasks of topic categorization, tone transition, and social message detection in advertisements. We use human experts to annotate the presence of underlying social messages and perceived tone for different segments of advertisement videos. For a broader content understanding, we also propose a condensed taxonomy of topics by combining information from multiple ads-specific expert sources. 
 We also investigate language-based reasoning through the use of large-language models on the ads transcripts. Furthermore, we show the utility of exploiting multiple modalities (audio, video and text) in a transformer-based fusion approach to obtain state-of-the-art results in the MM-AU benchmark tasks. Future directions would include: (a) expansion to additional benchmark tasks e.g.,  prediction of user intent, (b) understanding of causal mechanisms associated with tone transition and the (c) exploration of multimodal foundational models to tag and summarize ad videos with possible explanations. 
\begin{acks}
We would like to thank the Guggenheim Foundation for supporting the study.
\end{acks}


\bibliographystyle{ACM-Reference-Format}
\balance
\bibliography{sample-base}
\newpage
\appendix
\title{Supplementary: MM-AU: Towards multimodal understanding of advertisement videos}

\section{Introduction}
In this work, we provide the supplementary material associated with the submission: \textbf{MM-AU: Towards multimodal understanding of advertisement videos} 

\section{MM-AU dataset}
\subsection{Topic categorization:}
We provide the mapping between Cannes(CC) \cite{cannes-lions}, Ads of the World (AOW) \cite{AOW}  and Video-Ads (VA) \cite{Hussain2017AutomaticUO} coding schemes for obtaining the final set of topic categories as follows:

\begin{itemize}[leftmargin=*,labelsep=-\mylen]
    \item \textbf{Games:} Games and toys [\textcolor{blue}{\textbf{VA}}]; Gaming [\textcolor{red}{\textbf{AOW}}]
    \item \textbf{Household:} Household: Home Appliances, Furnishing [\textcolor{purple}{\textbf{CC}}]; Cleaning products, Home improvements and repairs, Home appliances [\textcolor{blue}{\textbf{VA}}]
    \item \textbf{Services:} Other services i.e. dating, tax, legal, loan, religious, printing, catering, etc. [\textcolor{blue}{\textbf{VA}}]; Professional Services [\textcolor{red}{\textbf{AOW}}].
    \item \textbf{Misc:} Miscellaneous, Business equipment and services [\textcolor{purple}{\textbf{CC}}]; Petfood, Political candidates (Politics) [\textcolor{blue}{\textbf{VA}}]; Pets [\textcolor{red}{\textbf{AOW}}]
    \item \textbf{Sports:} Sports equipment and activities [\textcolor{blue}{\textbf{VA}}]; Sports [\textcolor{red}{\textbf{AOW}}]
    \item \textbf{Banking:} Banking and services [\textcolor{purple}{\textbf{CC}}]; Financial services [\textcolor{blue}{\textbf{VA}}]; Finance [\textcolor{red}{\textbf{AOW}}]
    \item \textbf{Clothing:} Clothing, Footwear \& Accessories [\textcolor{purple}{\textbf{CC}}]; Clothing and accessories [\textcolor{blue}{\textbf{VA}}]; Personal Accessories [\textcolor{red}{\textbf{AOW}}] 
    \item \textbf{Industrial and agriculture:} Industrial, Agriculture Public Interest, Agriculture Professional Services [\textcolor{red}{\textbf{AOW}}]
    \item \textbf{Leisure:} Entertainment \& Leisure [\textcolor{purple}{\textbf{CC}}]; Gambling (lotteries, casinos, etc.) [\textcolor{blue}{\textbf{VA}}]; Recreation, Gambling [\textcolor{red}{\textbf{AOW}}]
    \item \textbf{Publications \& media:} Media \& Publications  [\textcolor{purple}{\textbf{CC}}]; Media and arts [\textcolor{blue}{\textbf{VA}}]; TV Promos, Music, Media, Movies [\textcolor{red}{\textbf{AOW}}]
    \item \textbf{Health:} Healthcare \& Pharmacy [\textcolor{purple}{\textbf{CC}}]; Health care and medications [\textcolor{blue}{\textbf{VA}}]; Health, Pharmaceutical [\textcolor{red}{\textbf{AOW}}]
    \item \textbf{Car:} Cars \& Automotive Products \& Services [\textcolor{purple}{\textbf{CC}}]; Car [\textcolor{blue}{\textbf{VA}}]; Automotive [\textcolor{red}{\textbf{AOW}}]
    \item \textbf{Electronics:} Home electronics and audio-visual [\textcolor{purple}{\textbf{CC}}]; Electronics, Phone, TV and internet service providers [\textcolor{blue}{\textbf{VA}}]; Electronics [\textcolor{red}{\textbf{AOW}}]
    \item \textbf{Cosmetics:} Cosmetics \& Toiletries [\textcolor{purple}{\textbf{CC}}]; Beauty products and cosmetics, Baby products [\textcolor{blue}{\textbf{VA}}]; Beauty [\textcolor{red}{\textbf{AOW}}]
    \item \textbf{Food and drink:} Savoury Foods, Sweet Foods \& Snacks, Non Alcoholic drinks, Alcoholic drinks [\textcolor{purple}{\textbf{CC}}]; Chocolate, Chips, Seasoning, Coffee, Soda, juice, milk, energy drinks, water, Alcohol [\textcolor{blue}{\textbf{VA}}]; Food, Non-Alcoholic Drinks, Confectionery, Alcoholic drinks [\textcolor{red}{\textbf{AOW}}]
    \item \textbf{Awareness:} Charities and non-profit [\textcolor{purple}{\textbf{CC}}]; Environment, Animal rights, Human rights, Safety, Smoking, Alcohol Abuse, Domestic Violence, Self-esteem, cyberbullying [\textcolor{blue}{\textbf{VA}}]; Education, Agency Self-Promo [\textcolor{red}{\textbf{AOW}}]
    \item \textbf{Travel and transport:} Travel \& Transport [\textcolor{purple}{\textbf{CC}}]; Vacation and travel [\textcolor{blue}{\textbf{VA}}]; Transport, Hospitality [\textcolor{red}{\textbf{AOW}}]
    \item \textbf{Retail:} Retail \& e-commerce [\textcolor{purple}{\textbf{CC}}]; Shopping (department stores, drug stores, groceries, etc.) [\textcolor{blue}{\textbf{VA}}]; Retail Services [\textcolor{red}{\textbf{AOW}}]
\end{itemize}
The taxonomy sources are listed within [.] for respective subcategories for the final list of topic categories. 

\subsection{Annotation Framework:}
\begin{figure*}[h!]
    \centering
    \includegraphics[width=\textwidth]{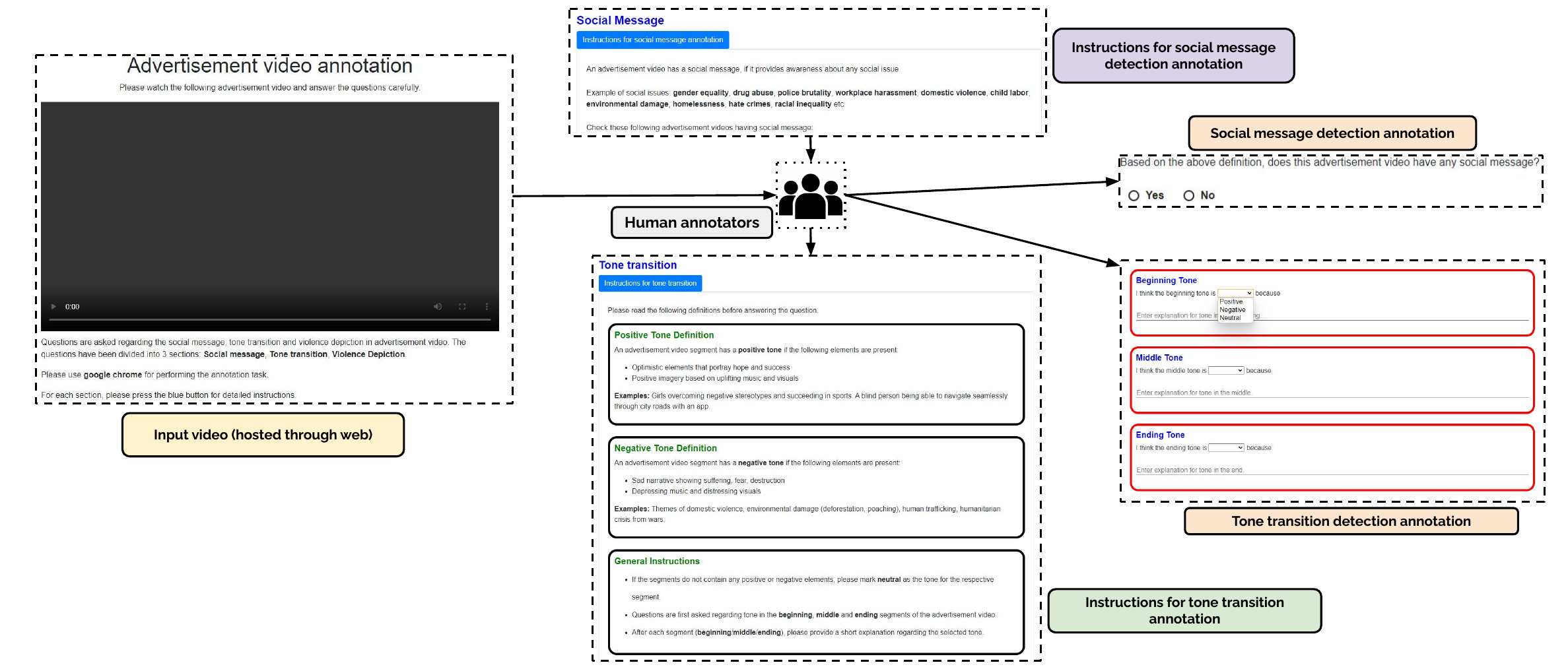}
    \caption{Outline of the annotation framework for tone transition and social message detection problem}
    \label{annot_framework}
\end{figure*}
\begin{figure}[h!]
    \centering
    \includegraphics[width=\columnwidth]{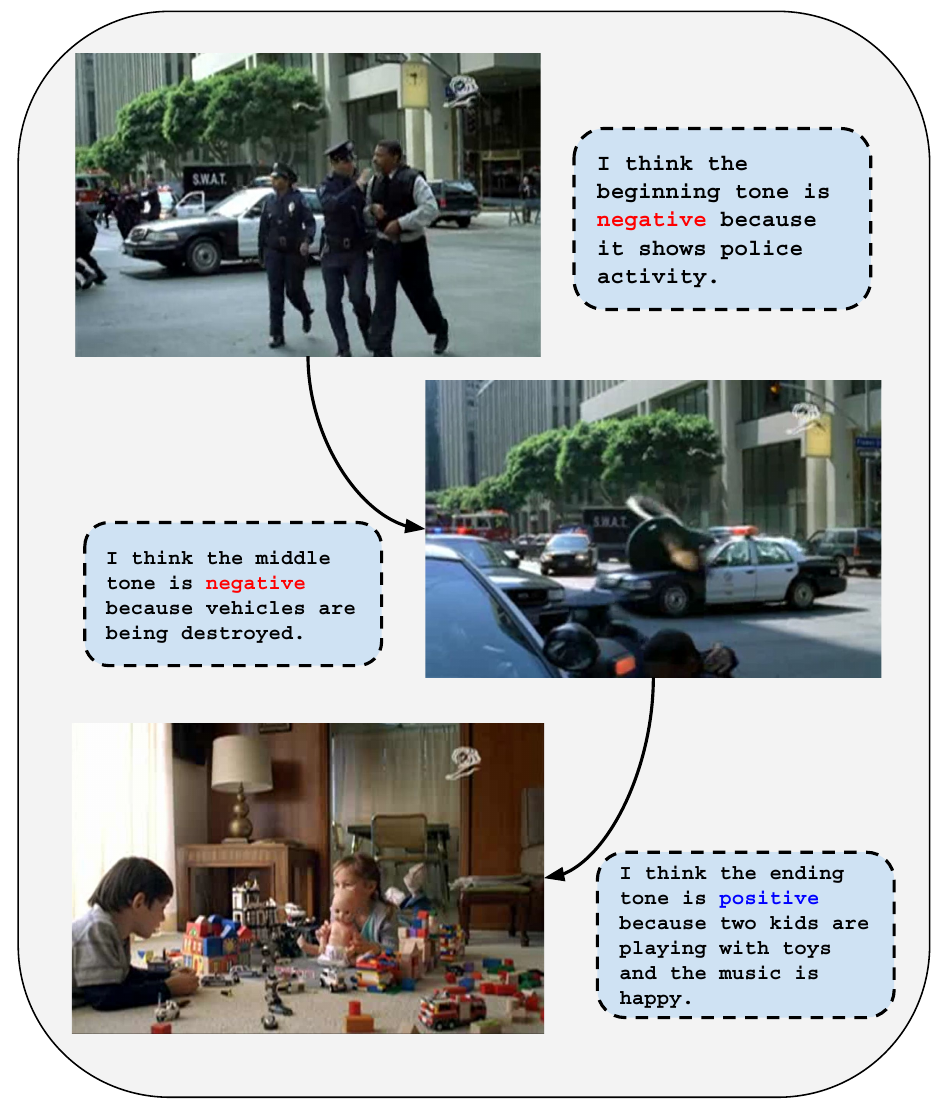}
    \caption{Example provided to the annotators showing the tone transition and associated explanations. Video source: \cite{cannes-lions}}
    \label{tone_transition}
\end{figure}
\begin{figure*}[h!]
\centering
    \includegraphics[width=0.8\textwidth]{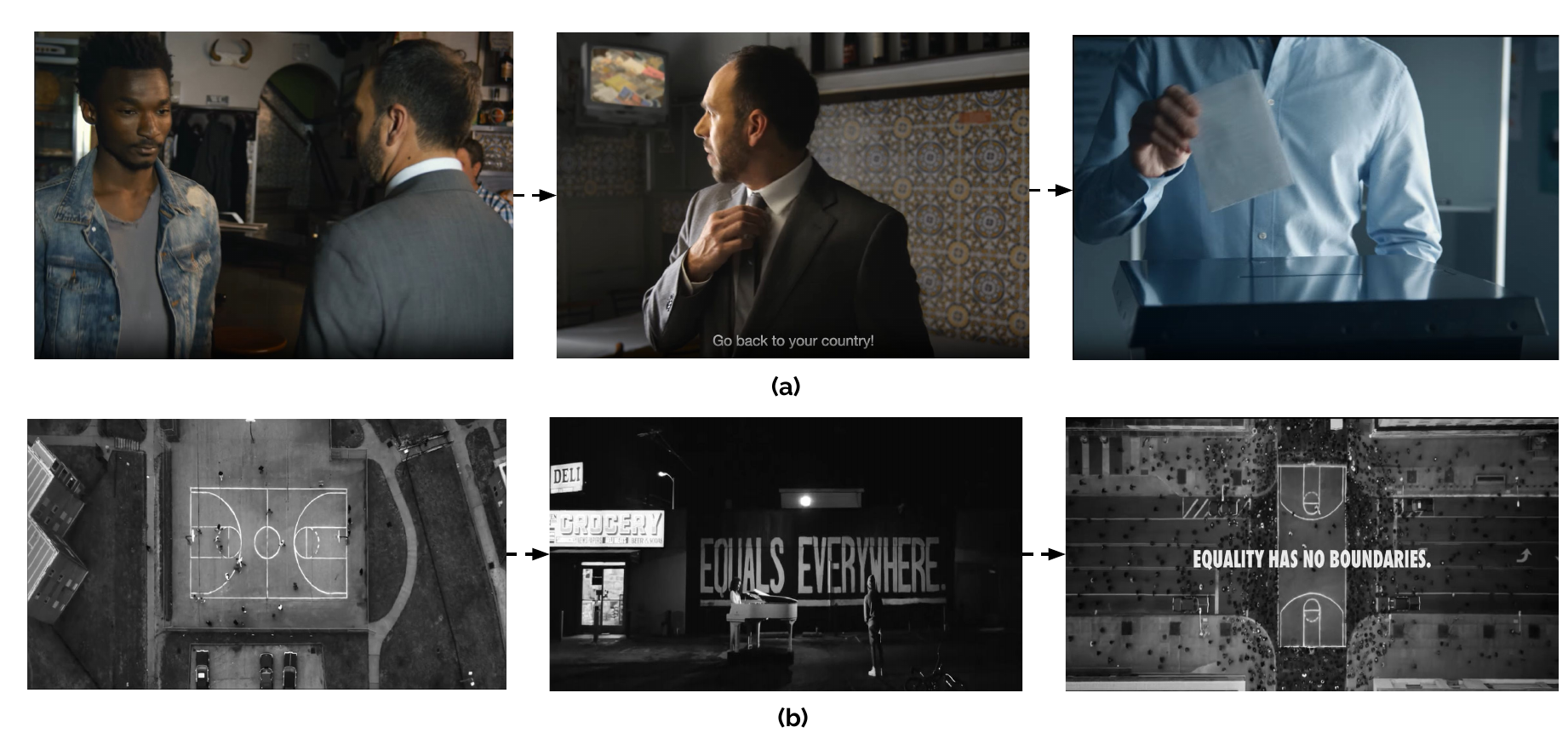}
    \caption{Example videos associated with absence/presence of social message provided to the annotators. (a) Frame transition associated with an example video urging people to vote (b) Frame transition associated with an example video emphasizing equal opportunities for everyone in sports. Video source: \cite{AOW}}
    \label{social_message}
\end{figure*}

In Fig \ref{annot_framework}, we show the outline of the framework provided to the annotators for marking the tone associated with the start, middle, and end segments and the absence/presence of a social message. The framework is hosted on Amazon Mechanical Turk \footnote{https://www.mturk.com/} platform. For marking the perceived tone labels associated with start, middle, and end segments, the annotators are asked to read the definitions of positive and negative tone and accompanying general instructions. For marking the respective tone labels, the annotators select options from the provided drop-down menus and provide general explanations for the respective segments. We provide a sample example to the annotators regarding tone transition and associated explanations, as shown in Fig \ref{tone_transition}. As seen in Fig \ref{tone_transition}, the beginning (start) segment has a perceived \textcolor{red}{\textbf{negative}} tone because police activity is being shown. The middle segment also shows a \textcolor{red}{\textbf{negative}} tone because vehicles are being destroyed, followed by a \textcolor{blue}{\textbf{positive}} tone at the ending portion because the kids are playing with toys. Further, annotators are also provided with example videos showing different forms of social messages. In Fig \ref{social_message} (a), frame transitions are shown from an example video urging everyone to vote since voters having bias can cast votes in their absence. In Fig  \ref{social_message} (b), sample frames in the sequence are shown from another example video highlighting the importance of equal opportunities for everyone in sports.
\subsection{Dataset analysis:}
\subsubsection{Agreement distribution:}
\begin{figure*}
\begin{subfigure}{.33\textwidth}
  \centering
  \includegraphics[width=\linewidth]{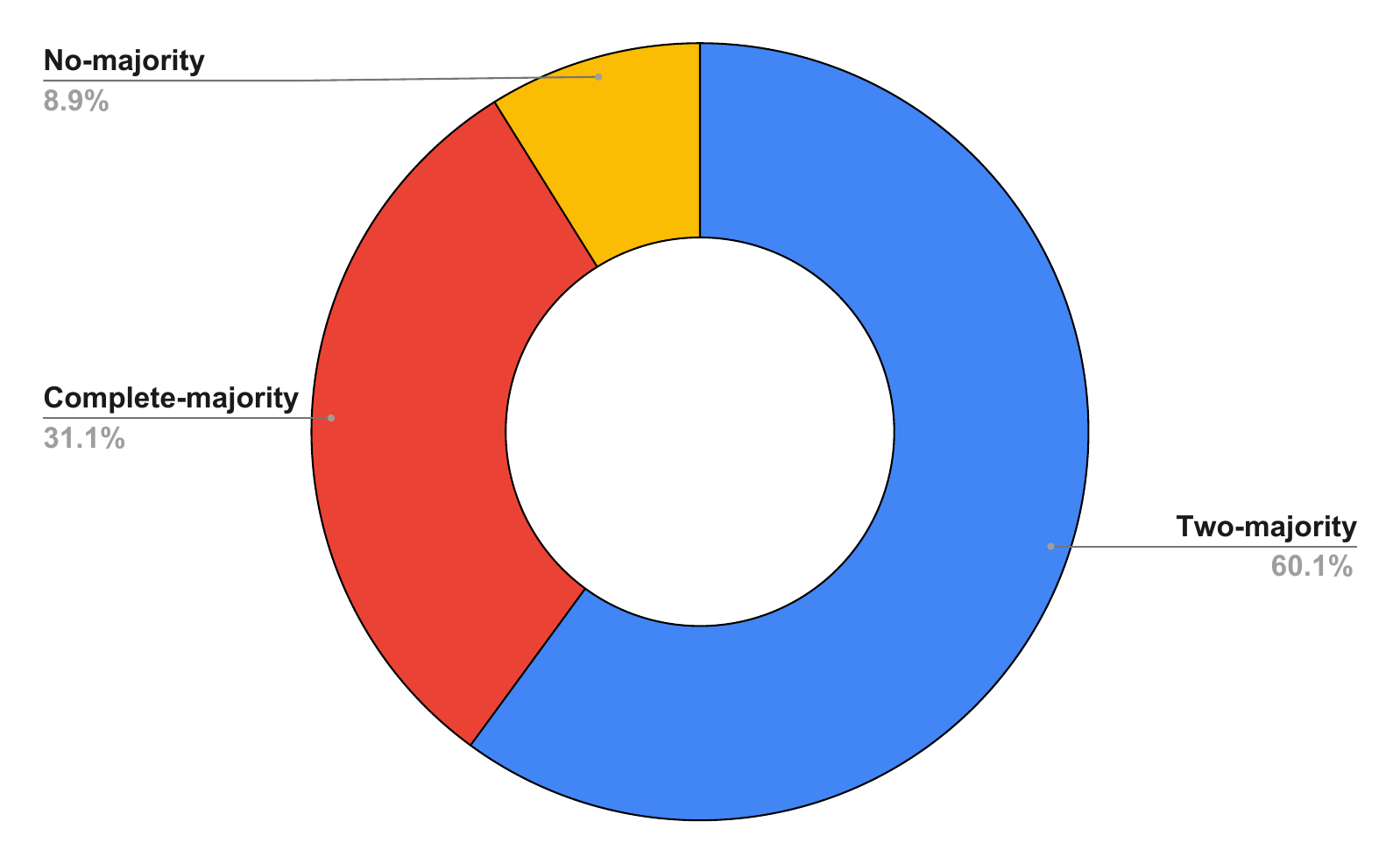}
  \caption{Start segment}
  \label{start_tone}
\end{subfigure}%
\begin{subfigure}{.33\textwidth}
  \centering
  \includegraphics[width=\linewidth]{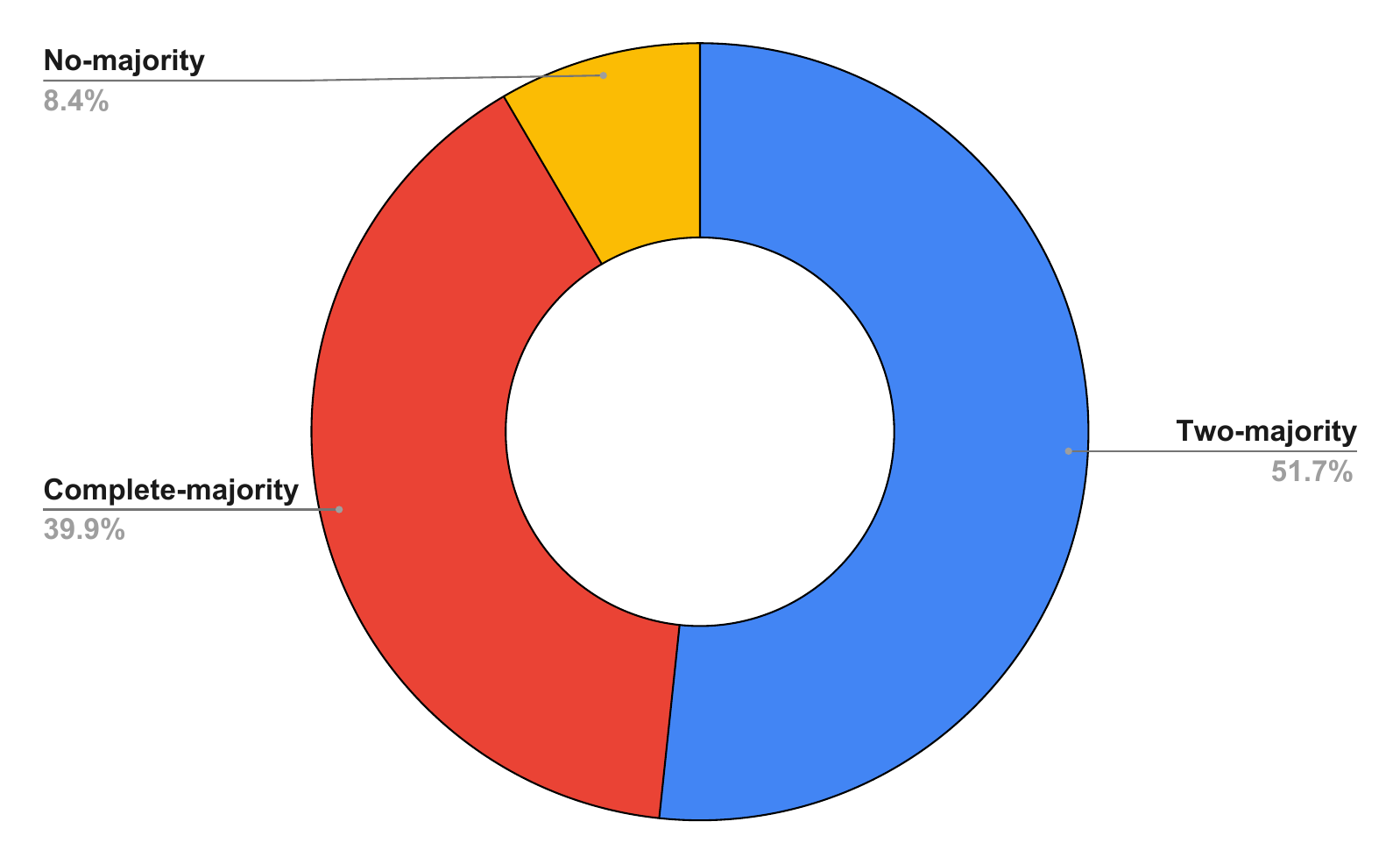}
  \caption{Middle segment}
  \label{mid_tone}
\end{subfigure}
\begin{subfigure}{.33\textwidth}
  \centering
  \includegraphics[width=\linewidth]{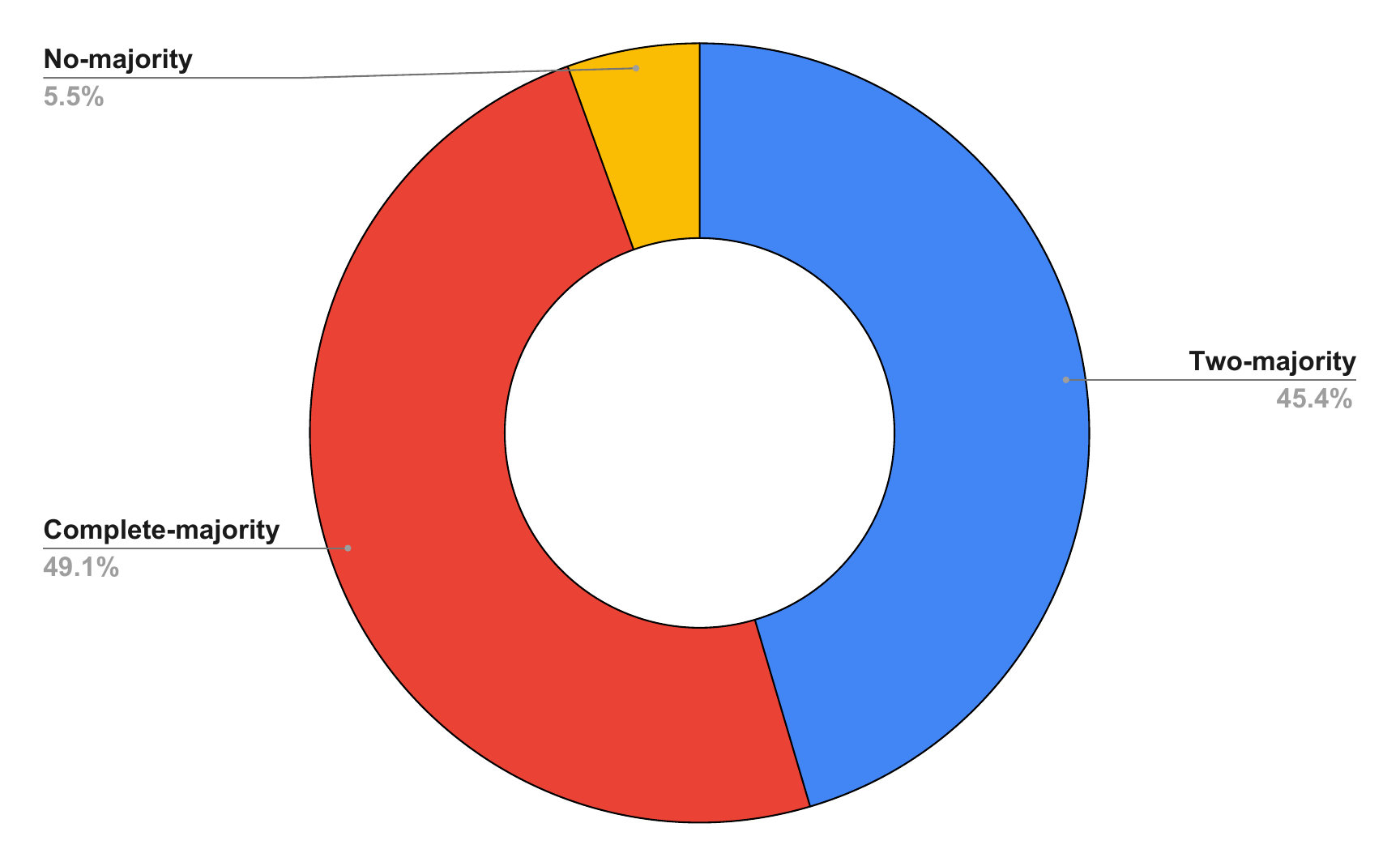}
  \caption{End segment}
  \label{end_tone}
\end{subfigure}
\caption{Distribution of annotator agreement (among 3 annotators) across (a) start, (b) middle, and (c) ending segments in MM-AU dataset (8399 videos) }
\label{start_mid_end_annot_agreement}
\end{figure*}

 We present the distribution of agreement among the annotators providing the final annotations for social message and tone transition detection tasks in Figure \ref{start_mid_end_annot_agreement}. We define the types of agreements as follows:
 \begin{itemize}[leftmargin=*,labelsep=-\mylen]
     \item \textbf{\underline{No-majority}:} No agreement exists between 3 annotators
     \item \textbf{\underline{Two-majority}:} Agreement exists between 2 annotators (out of 3 annotators)
     \item \textbf{\underline{Majority}:} Agreement exists between 3 annotators.
 \end{itemize}
 From Fig \ref{start_mid_end_annot_agreement} (a), we can see that the agreements among annotators for the start segment are distributed as: \textbf{Two-majority: 60.1\%}, \textbf{Complete-majority: 31.1\%}, \textbf{No-majority: 8.9\%}. As seen in Fig \ref{start_mid_end_annot_agreement} (b), The share of the complete majority agreement increases to \textbf{39.9\%} for the middle segment of the video. Since the ending of advertisement videos is predominantly positive, we obtain higher agreement values in terms of complete majority (\textbf{49.1\%}) and lower values for no majority (\textbf{5.5\%}) as compared to start and middle segments.
\subsubsection{Social message intersection:}
\begin{figure*}
\begin{subfigure}{.50\textwidth}
  \centering
  \includegraphics[width=\linewidth]{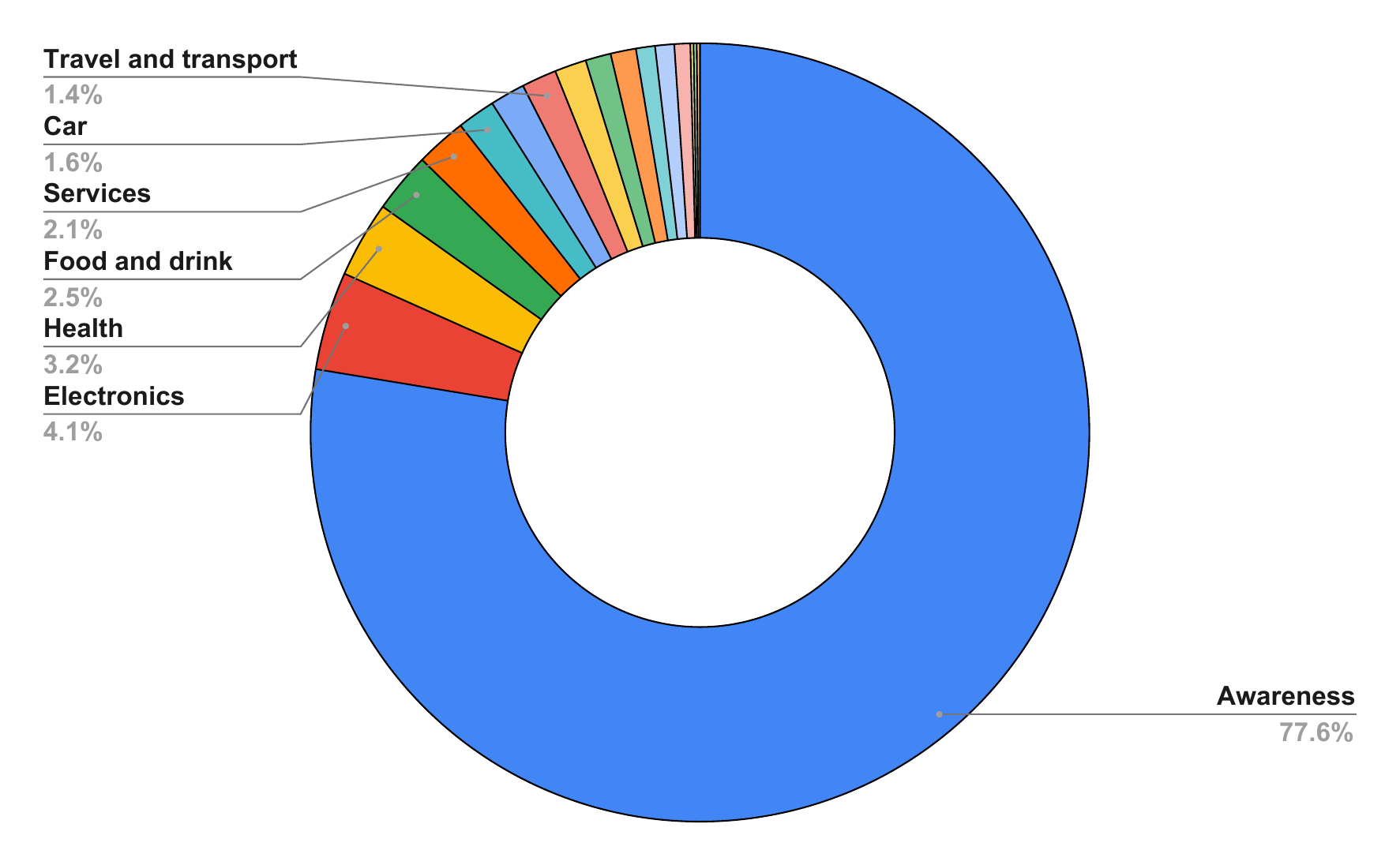}
  \caption{Distribution of Topics wrt videos having social message}
  \label{topic_social_message}
\end{subfigure}%
\begin{subfigure}{.50\textwidth}
  \centering
  \includegraphics[width=\linewidth]{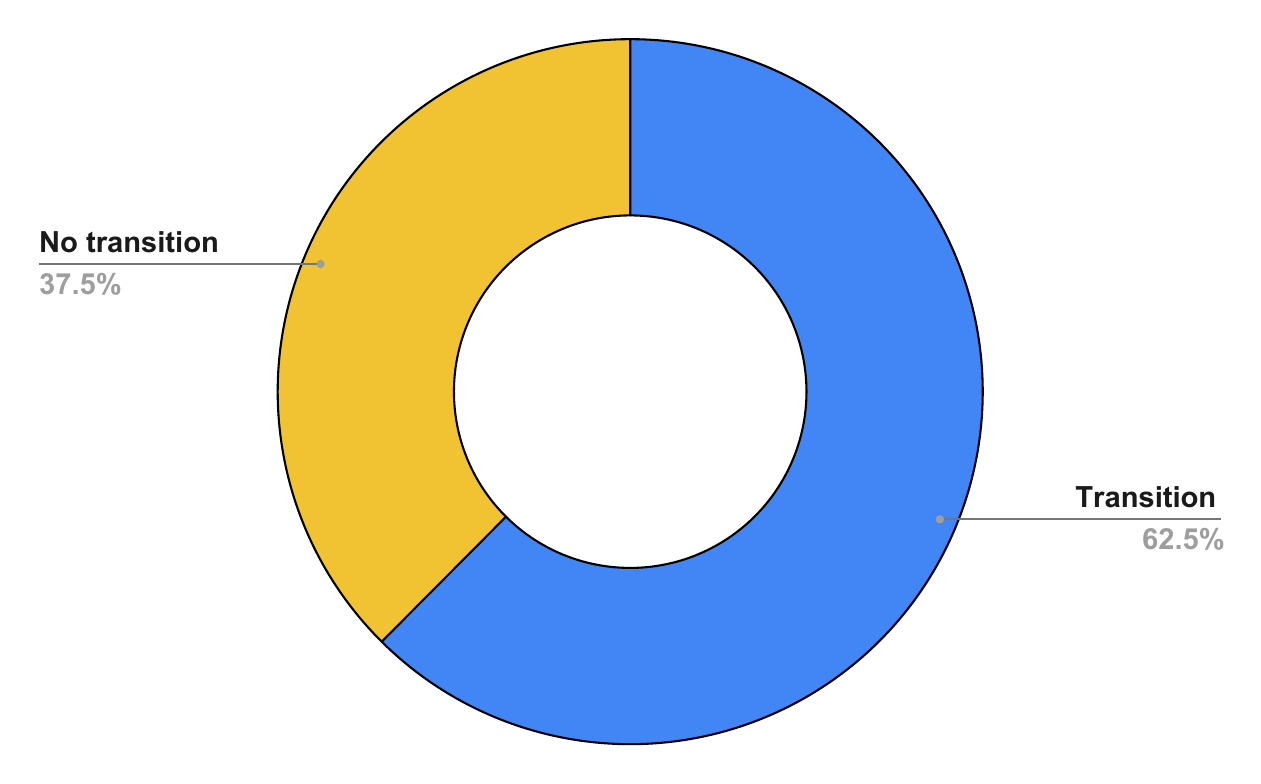}
  \caption{Distribution of Tone transition wrt videos having social message}
  \label{tone_transition_social_message}
\end{subfigure}
\caption{Distribution of topics and perceived tone transition across videos having the presence of social message (739 videos) in MM-AU dataset}
\label{topic_tone_transition_social_message}
\end{figure*}

\begin{figure*}
\begin{subfigure}{.33\textwidth}
  \centering
  \includegraphics[width=\linewidth]{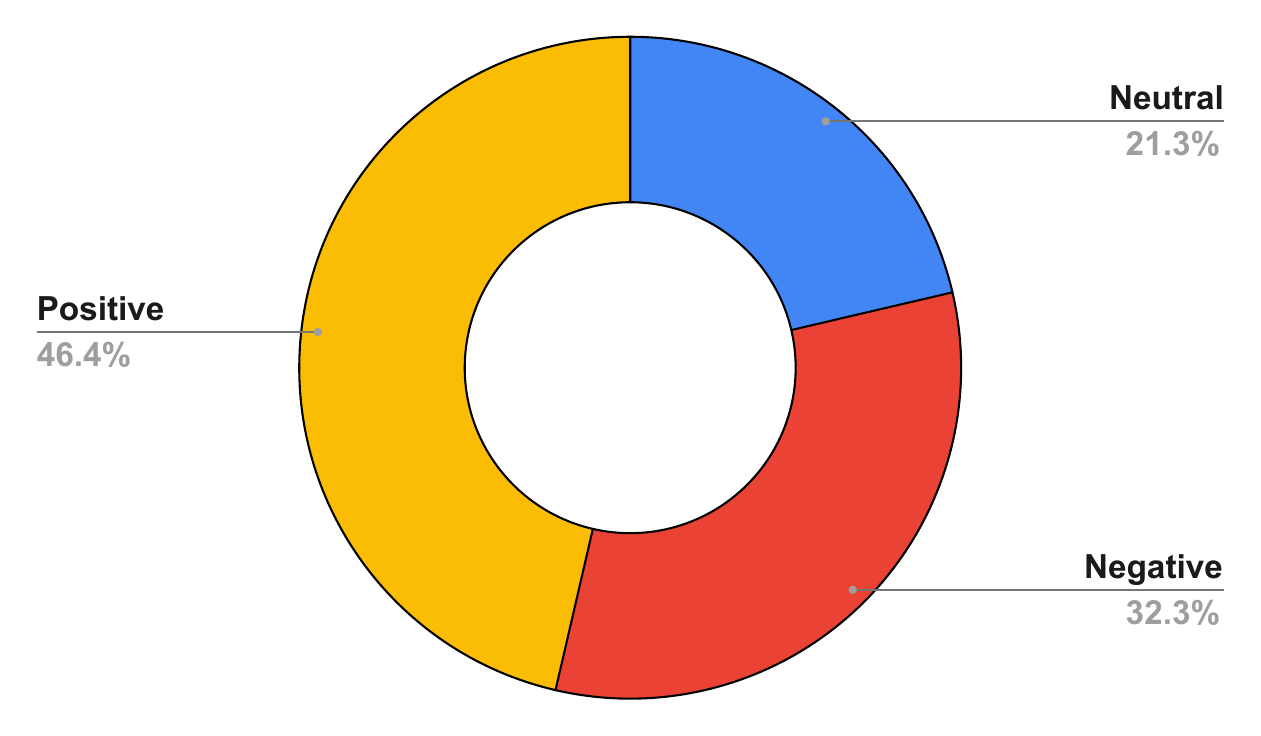}
  \caption{Start segment}
  \label{start_tone_social_message}
\end{subfigure}%
\begin{subfigure}{.33\textwidth}
  \centering
  \includegraphics[width=\linewidth]{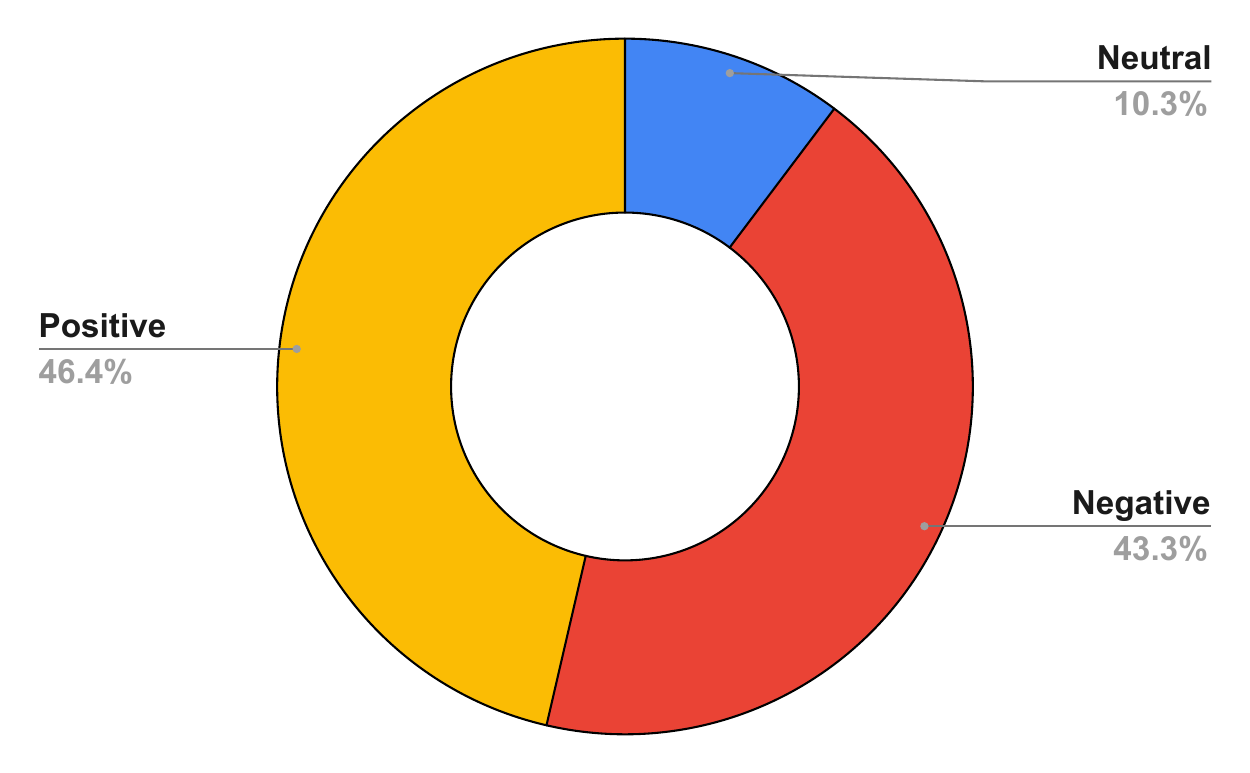}
  \caption{Middle segment}
  \label{mid_tone_social_message}
\end{subfigure}
\begin{subfigure}{.33\textwidth}
  \centering
  \includegraphics[width=\linewidth]{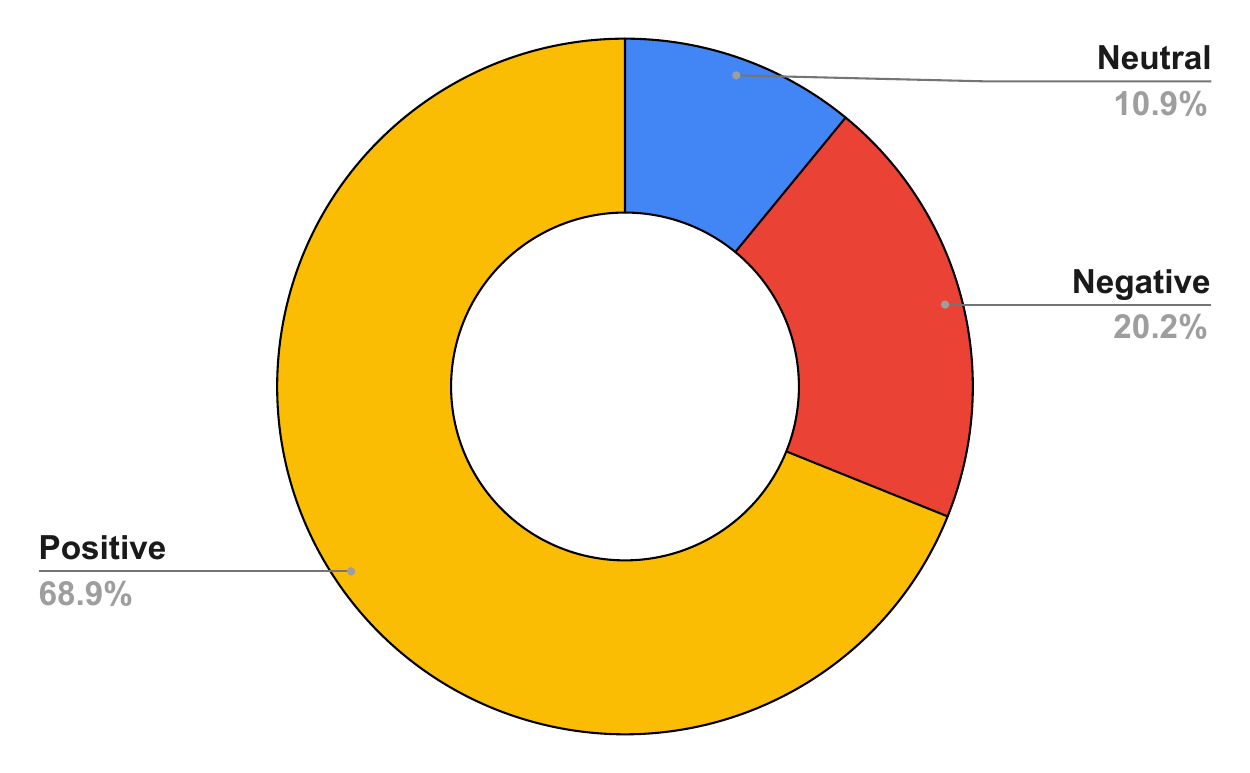}
  \caption{End segment}
  \label{end_tone_social_message}
\end{subfigure}
\caption{Distribution of perceived tone labels across the (a) start, (b) middle, and (c) ending segments in MM-AU dataset for videos having social message (739 videos) }
\label{start_mid_end_soc_msg_tone}
\end{figure*}

\begin{figure*}
\begin{subfigure}{.50\textwidth}
  \centering
  \includegraphics[width=\linewidth]{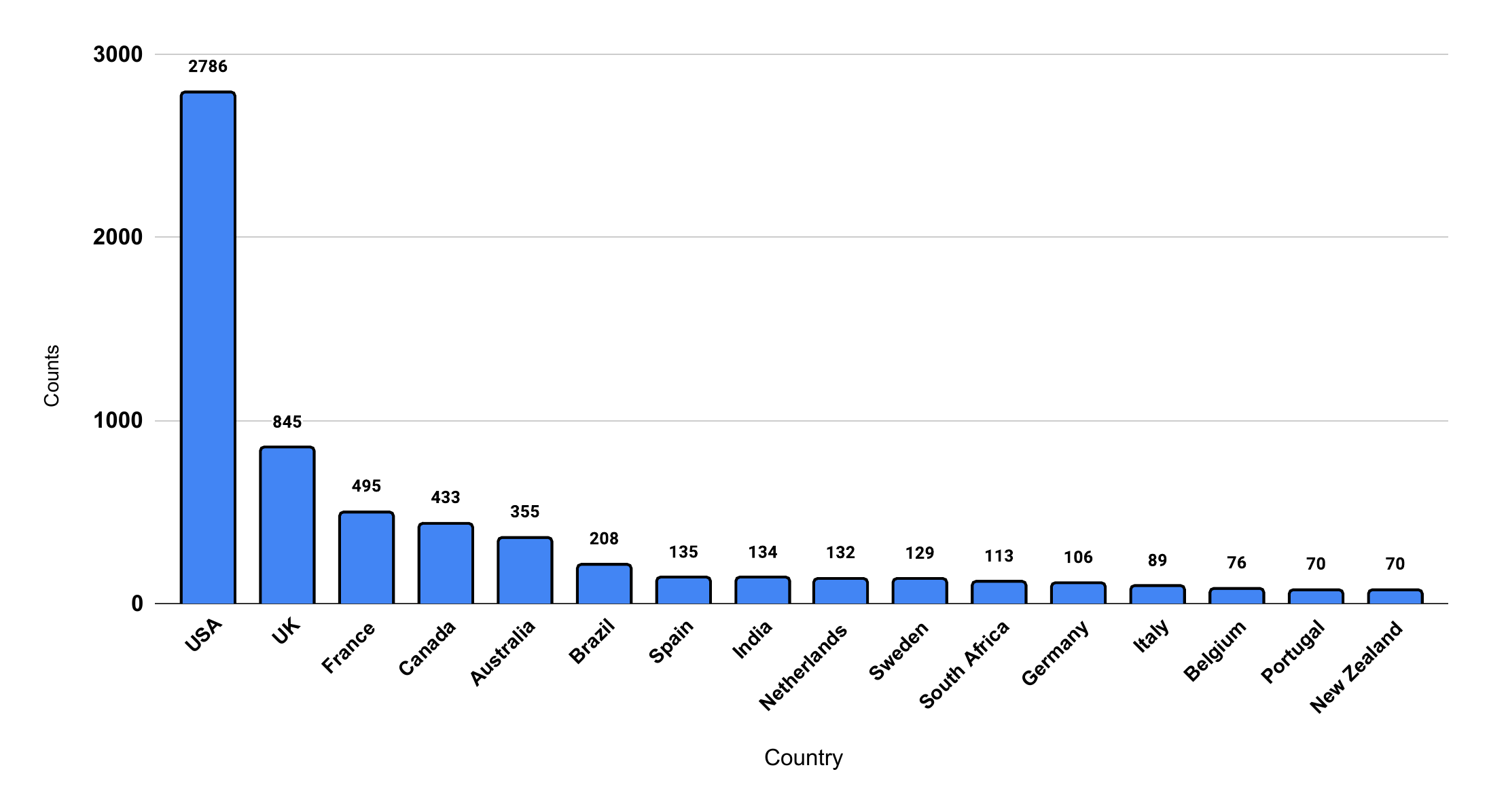}
  \caption{Country-wise distribution of videos in MM-AU dataset}
  \label{country_wise_videos_list}
\end{subfigure}%
\begin{subfigure}{.50\textwidth}
  \centering
  \includegraphics[width=\linewidth]{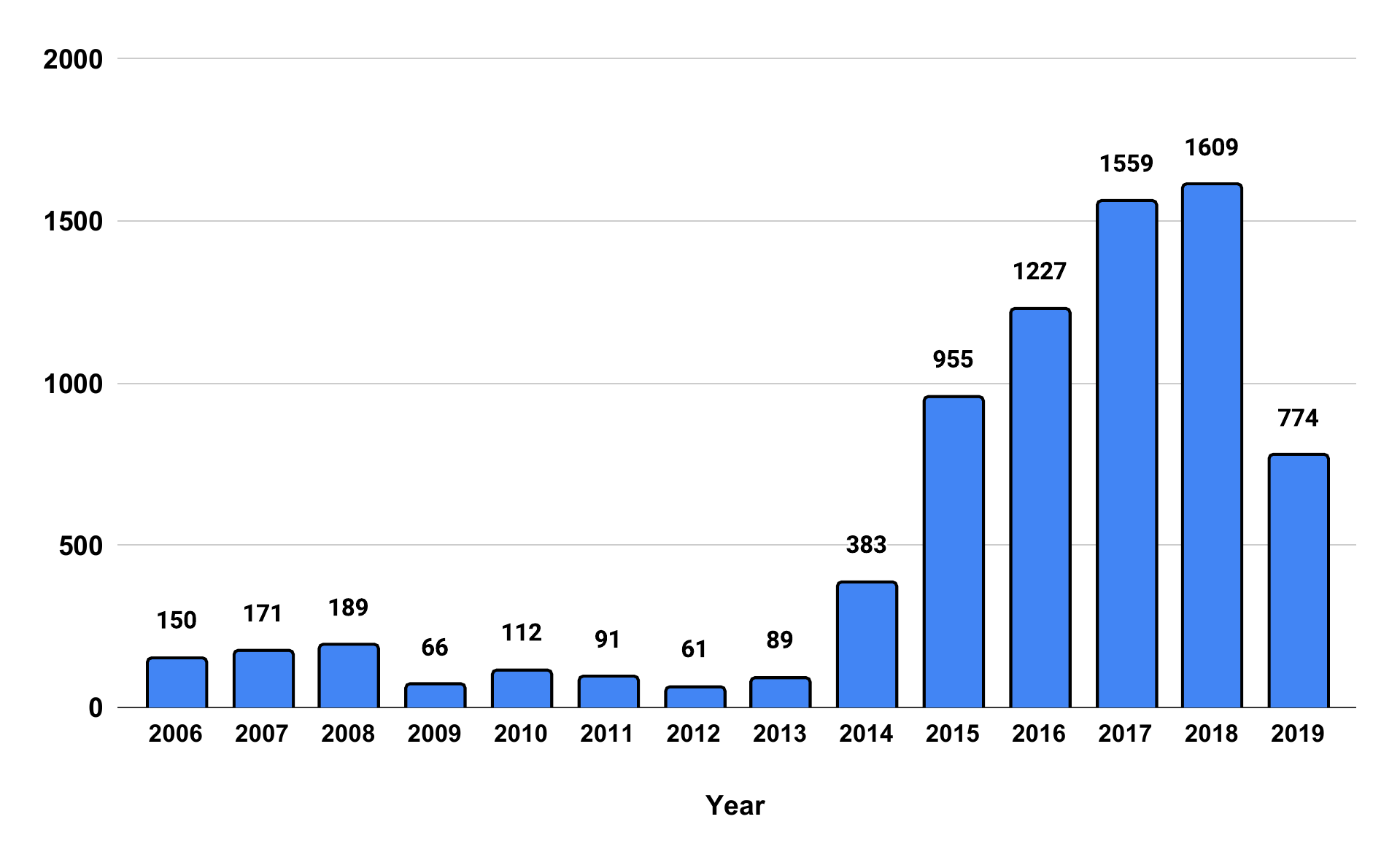}
  \caption{Year-wise distribution of videos in MM-AU dataset}
  \label{year_wise_videos_list}
\end{subfigure}
\caption{Distribution of country and year across videos in MM-AU dataset. Distribution here is shown for 7436 videos (out of 8399 videos in MM-AU dataset)}
\label{country_year_wise_video_list}
\end{figure*}

\begin{figure*}[h!]
    \centering
    \includegraphics[width=\textwidth]{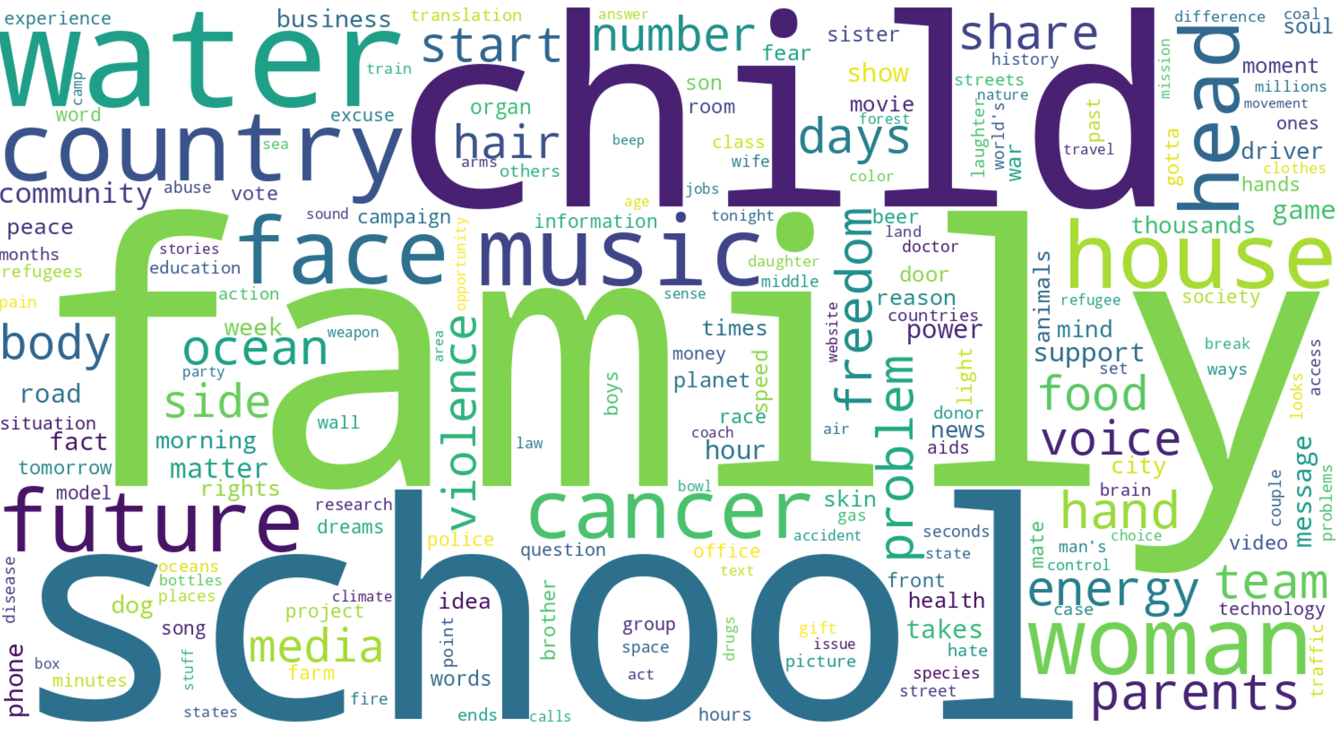}
    \caption{Word cloud of important terms from ads transcripts extracted using Whisper large \cite{Radford2022RobustSR} for videos having social message (742 out of 8399 videos). The size of the concepts(words) is proportional to the frequency of occurrence.}
    \label{soc_msg_word_cloud}
\end{figure*}

 We further explore the intersection between the presence of social message and associated topics and tone transition (including tone labels for different segments of the video). From Fig \ref{topic_tone_transition_social_message} (a), we can see that out of 739 videos having social messages, \textbf{77.9\%} have \textbf{Awareness} as the associated topic label. The \textbf{Awareness} topic label includes subcategories related to social issues involving environmental damage, animal rights, smoking, alcohol abuse, domestic violence, refugee crisis, cyberbullying, etc. Further, as shown in Fig \ref{topic_tone_transition_social_message} (b), we observe a greater incidence of transitions in perceived tone (\textbf{62.5\%}) associated with videos having social messages. A detailed breakdown of segment-wise perceived tone labels for videos having social messages is shown in Fig \ref{start_mid_end_soc_msg_tone}. We can see an increase in the perceived negative tone from \textbf{32.3\%} to \textbf{43.3\%}. This can be attributed to the narrative structure of the advertisement videos having the presence of a social message where the middle segment primarily portrays negative elements i.e. environmental damage, human suffering etc to set up the conclusion. We also extract transcripts from ads videos using the multilingual Whisper large model \cite{Radford2022RobustSR}. Based on the contents of the transcripts, we extract concepts relevant to social issues and visualize them in Fig \ref{soc_msg_word_cloud}. We can see certain words relevant to social issues like \textbf{diseases} (\textit{cancer}), \textbf{environment} (\textit{ocean}, \textit{bottles}, \textit{water}, \textit{energy}, \textit{coal}),  \textbf{conflict} (\textit{war}, \textit{violence}, \textit{refugees}).
\subsubsection{Metadata distribution:}
\begin{figure*}[h!]
    \centering
    \includegraphics[width=0.7\textwidth]{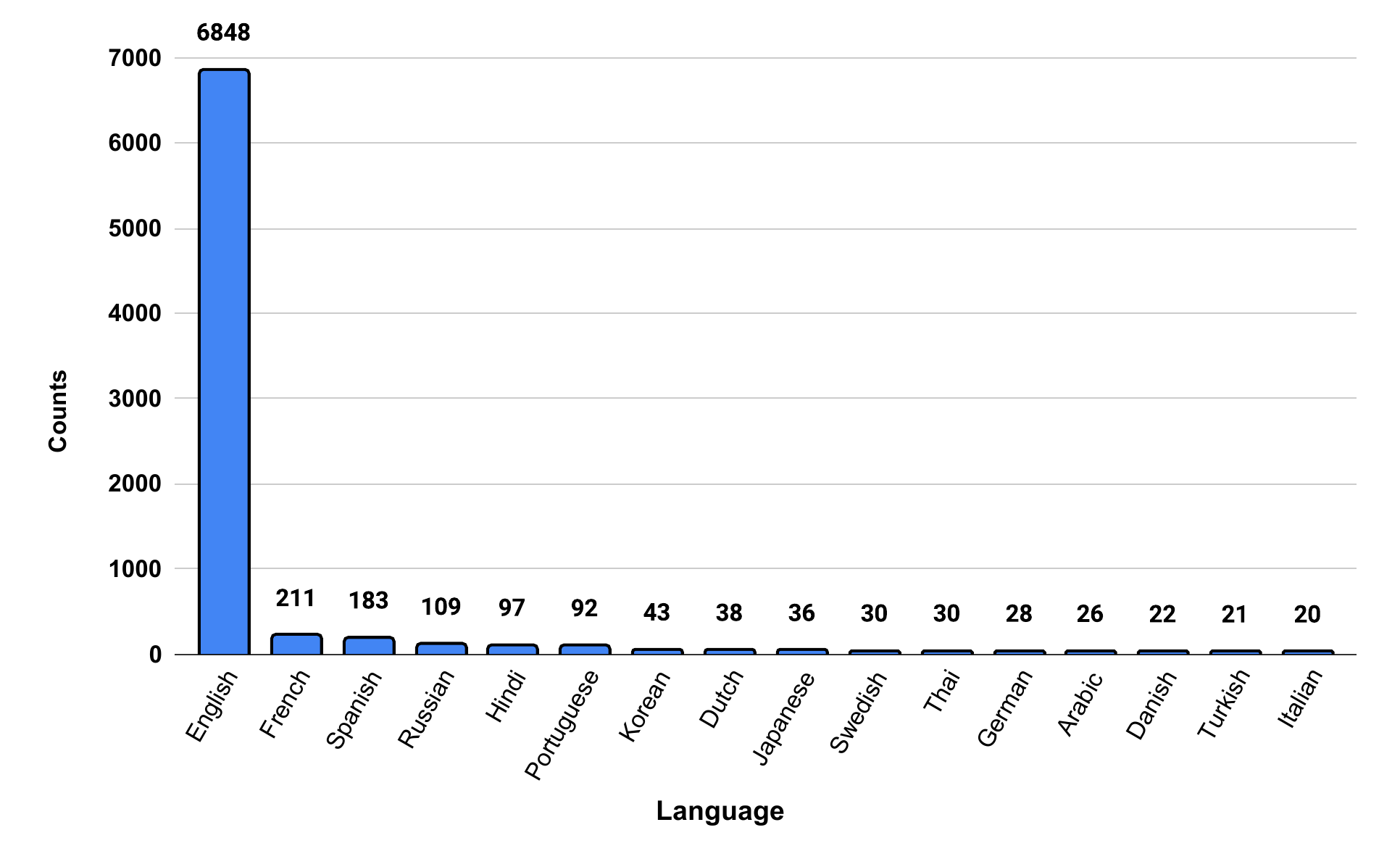}
    \caption{Language distribution in the transcripts extracted using Whisper large \cite{Radford2022RobustSR} for MM-AU dataset. Languages having less than equal to 20 occurrences are shown.}
    \label{whisper_mm_au_dataset}
\end{figure*}

In Fig \ref{country_year_wise_video_list} (a), we show the distribution of the 99 countries in MM-AU dataset. We consider only those videos from Ads of the World (AOW) and the Cannes-Lions archive i.e. (7436 out of 8399 videos in MM-AU dataset), for which metadata information is available. Since metadata with Video-Ads dataset \cite{Hussain2017AutomaticUO} is not publicly available, we don't use those videos for visualizing the country and year-wise information. We can see from Fig \ref{country_year_wise_video_list} (a) that a major chunk of videos is from \textbf{USA}(2786), \textbf{UK}(845), \textbf{France}(495), \textbf{Canada}(433). For the purpose of visualization, we show only those countries that have at least 70 videos. From \ref{country_year_wise_video_list} (b), we can see that a significant chunk of videos i.e. \textbf{6507} (out of 7436 videos), lie in the year range 2014-2019. From \ref{whisper_mm_au_dataset}, we can see that English is the dominant language (\textbf{6848}) followed by French (\textbf{211}), Spanish (\textbf{183}) in MM-AU dataset.

\subsubsection{Explanation distribution:} 

From Table \ref{segment table}, we can see that the lengths of the explanations provided by the annotators for the start, middle, and ending segments are fairly distributed, with average lengths varying from 7.583 (\textbf{Start}) to 7.435 (\textbf{Ending}).
\begin{table}[h!]
\begin{tabular}{@{}cl@{}}
\toprule
\multicolumn{1}{l}{Segment} & Length \\ \midrule
Start                       & 7.583 ± 5.263 \\
Middle                      & 7.857 ± 5.325\\
Ending                      & 7.435 ± 5.035\\
\end{tabular}
\caption{Distribution of segment-wise length of explanations}
\end{table}
\label{segment table}
\section{Experiments}

\subsection{Language based reasoning}

We investigate the zero-shot performance of several large language models i.e. \texttt{GPT-4}\cite{OpenAI2023GPT4TR}, \texttt{Opt-IML} \cite{Iyer2022OPTIMLSL}, \texttt{Flan-T5} (XXL,XL,L) \cite{Chung2022ScalingIL} and \texttt{Alpaca} \cite{alpaca} on the benchmark tasks associated with \textbf{MM-AU} dataset. For zero-shot evaluation, we report the results on 1670 non-empty transcripts out of the test split of 1692 samples.
\subsubsection{Flan-T5:}
For \texttt{Flan-T5}, we use the following prompts for the social message (\textbf{SM}), tone transition (\textbf{TT}), topic categorization(\textbf{Topic}) tasks: 
\begin{itemize}[leftmargin=*,labelsep=-\mylen]

\item \textbf{\underline{TT:}} 
\texttt{<Text from transcript>} \\
\textit{Based on the given text transcript from the advertisement, determine if the advertisement has any transitions in tones.} \\ 
\textbf{OPTIONS:}
\begin{itemize}
\item[-] Transition
\item[-] No transition
\end{itemize}
\textbf{ANSWER:}

\item \textbf{\underline{SM:}}
\texttt{<Text from transcript>} \\
\textit{An advertisement video has a social message if it provides awareness about any social issue. Examples of social issues: gender equality, drug abuse, police brutality, workplace harassment, domestic violence, child labor, environmental damage, homelessness, hate crimes, racial inequality etc. Based on the given text transcript, determine if the advertisement has any social message.}\\
\textbf{OPTIONS:}
\begin{itemize}
\item[-] Yes
\item[-] No
\end{itemize}
ANSWER: 

\item \textbf{\underline{Topic:}}
\texttt{<Text from transcript>} \\
\textit{Associate a single topic label with the transcript from the given set:} \\
\textbf{OPTIONS:}
\begin{itemize}
\item [-] Games
\item [-] Household
\item [-] Services
\item [-] Sports 
\item [-] Banking 
\item [-] Clothing 
\item [-] Industrial and agriculture 
\item [-] Leisure 
\item [-] Publications media 
\item [-] Health 
\item [-] Car 
\item [-] Electronics 
\item [-] Cosmetics 
\item [-] Food and drink 
\item [-] Awareness 
\item [-] Travel and transport 
\item [-] Retail 
\end{itemize}
\textbf{ANSWER:}
\end{itemize}

\subsubsection{OPT:} For \texttt{OPT}, we use the following prompt templates for different tasks:

\begin{itemize}[leftmargin=*,labelsep=-\mylen]

    \item \textbf{\underline{TT:}} \textit{Instruction: In this task, you are given a transcription of an advertisement, determine if the advertisement has any transitions in tones.} \\
    \textbf{Transcription:} \texttt{<Text from transcript>}\\
    \textbf{OPTIONS:}
    \begin{itemize}
    \item[-] Transition
    \item[-] No transition
    \end{itemize}
   \textbf{Answer:}

    \item \textbf{\underline{SM:}} \textit{In this task, you are given a transcription of an advertisement. An advertisement video has a social message if it provides awareness about any social issue. Example of social issues: gender equality, drug abuse, police brutality, workplace harassment, domestic violence, child labor, environmental damage, homelessness, hate crimes, racial inequality etc. Your task is to give label "Yes" if the advertisement given has any social message, otherwise give label "No".} \\
    \textbf{Transcription:} \texttt{<Text from transcript>}\\
    \textbf{Answer:}
    
    \item \textbf{\underline{Topic:}} \textit{In this task, you are given a transcription of an advertisement. Your task is to associate a single topic label with the transcript from the given set.} \\
    \textbf{Transcription:} \texttt{<Text from transcript>}\\
    \textbf{OPTIONS:}
    \begin{itemize}
    \item [-] Games
    \item [-] Household
    \item [-] Services
    \item [-] Sports 
    \item [-] Banking 
    \item [-] Clothing 
    \item [-] Industrial and agriculture 
    \item [-] Leisure 
    \item [-] Publications media 
    \item [-] Health 
    \item [-] Car 
    \item [-] Electronics 
    \item [-] Cosmetics 
    \item [-] Food and drink 
    \item [-] Awareness 
    \item [-] Travel and transport 
    \item [-] Retail 
    \end{itemize}
    \textbf{Answer:}
\end{itemize}

\subsubsection{alpaca:} 
For \texttt{alpaca}, we use the following prompt templates for different tasks:

\begin{itemize}[leftmargin=*,labelsep=-\mylen]
    \item \textbf{\underline{TT:}} \textit{Instruction: In this task, you are given a transcription of an advertisement determine if the advertisement has any transitions in tones.}\\
\textbf{Transcription}: \texttt{<Text from transcript>}\\
\textbf{Options:}
\begin{itemize}
\item[-] Transition
\item[-] No transition
\end{itemize}
\textbf{Answer:}
\item \textbf{\underline{SM:}} \textit{Instruction: In this task, you are given a transcription of an advertisement. An advertisement video has a social message if it provides awareness about any social issue. Example of social issues: gender equality, drug abuse, police brutality, workplace harassment, domestic violence, child labor, environmental damage, homelessness, hate crimes, racial inequality etc. Based on the given text transcript, determine if the advertisement has any social message. }\\
    \textbf{Transcription:} \texttt{<Text from transcript>}\\
\textbf{Options:}
\begin{itemize}
\item[-] Yes
\item[-] No
\end{itemize}
\textbf{Answer:}
\item \textbf{\underline{Topic:}} \textit{Instruction: In this task, you are given a transcription of an advertisement. Your task is to associate a single topic label with the transcript from the given set.}\\
\textbf{Transcription:} \texttt{<Text from transcript>}\\
\textbf{Options:}
    \begin{itemize}
    \item [-] Games
    \item [-] Household
    \item [-] Services
    \item [-] Sports 
    \item [-] Banking 
    \item [-] Clothing 
    \item [-] Industrial and agriculture 
    \item [-] Leisure 
    \item [-] Publications media 
    \item [-] Health 
    \item [-] Car 
    \item [-] Electronics 
    \item [-] Cosmetics 
    \item [-] Food and drink 
    \item [-] Awareness 
    \item [-] Travel and transport 
    \item [-] Retail 
\end{itemize}
\textbf{Answer:}
\end{itemize}
For \texttt{GPT-4} we use the recently released API to pass the prompts for individual tasks.
For \texttt{Flan-T5} and \texttt{Opt-IML}, we use the publicly available models as a part of Huggingface \cite{Wolf2019HuggingFacesTS} library. For \texttt{alpaca}, we use the publicly available implementation in Github \cite{alpaca}. The large language models sometimes assign a label to the prediction that does not lie within the set of valid labels for the respective tasks. In the case of those samples, we randomly assign a label from the task-specific label taxonomy. 

\subsection{Unimodal and multimodal baselines}

\subsubsection{Unimodal baselines}
For the unimodal baselines based on \textbf{MHA} \cite{Transformers}, \textbf{LSTM} \cite{Hochreiter1997LongSM}, we use the following hyperparameter choices based on the input modalities:\\

\textbf{LSTM(Visual):} For LSTM with visual modality (shot-wise representations) for all the tasks, we use the following hyperparameter settings:
\begin{itemize}
    \item \textbf{Batch size:} 16
    \item \textbf{Optimizer:} \texttt{Adam} \cite{kingma2014adam} (lr=1e-4)
    \item \textbf{Maximum sequence length}: 35 (Shots)
    \item \textbf{Max epochs:} 50
    \item \textbf{Patience:} 5
    \item \textbf{Hidden size:} 256
    \item \textbf{Number of layers:} 2
\end{itemize}
\vspace{3mm}
\textbf{MHA(Visual)}: For MHA model with visual modality as input (shot-wise representations) for the social message and tone transition tasks, we use the following hyperparameter settings:
\begin{itemize}
    \item \textbf{Batch size:} 16
    \item \textbf{Optimizer:} \texttt{Adam} \cite{kingma2014adam} (lr=1e-5)
    \item \textbf{Maximum video sequence length}: 35 (Shots)
    \item \textbf{Max epochs:} 50
    \item \textbf{Patience:} 5
    \item \textbf{Hidden size:} 256
    \item \textbf{Number of layers:} 4 
    \item \textbf{Number of heads:} 4
     \item \textbf{input\_dropout:} 0.2
  \item \textbf{output\_dropout:} 0.2
  \item \textbf{model\_dropout:} 0.2
\end{itemize}
For topic categorization, all the hyperparameters remain the same except the optimizer and learning rate i.e. \texttt{AdamW} \cite{Loshchilov2017DecoupledWD} (lr=1e-4).
\\\\
\textbf{MHA(Audio)}: 
 For MHA model with audio modality as input for the social message and tone transition tasks, we use the following hyperparameter settings:
\begin{itemize}
    \item \textbf{Batch size:} 16
    \item \textbf{Optimizer:} \texttt{Adam} \cite{kingma2014adam} (lr=1e-4)
    \item \textbf{Maximum audio sequence length}: 14
    \item \textbf{Hidden size:} 256
    \item \textbf{Number of layers:} 2
    \item \textbf{Number of heads:} 2
     \item \textbf{input\_dropout:} 0.2
  \item \textbf{output\_dropout:} 0.2
  \item \textbf{model\_dropout:} 0.2
\end{itemize}
For MHA model with audio modality as input for the topic categorization task, all the hyperparameters remain the same except the following:
\begin{itemize}
    \item \textbf{Number of layers:} 4
    \item \textbf{Number of heads:} 4
\end{itemize}
\begin{figure*}[h!]
\begin{subfigure}{.50\textwidth}
  \centering
  \includegraphics[width=\linewidth]{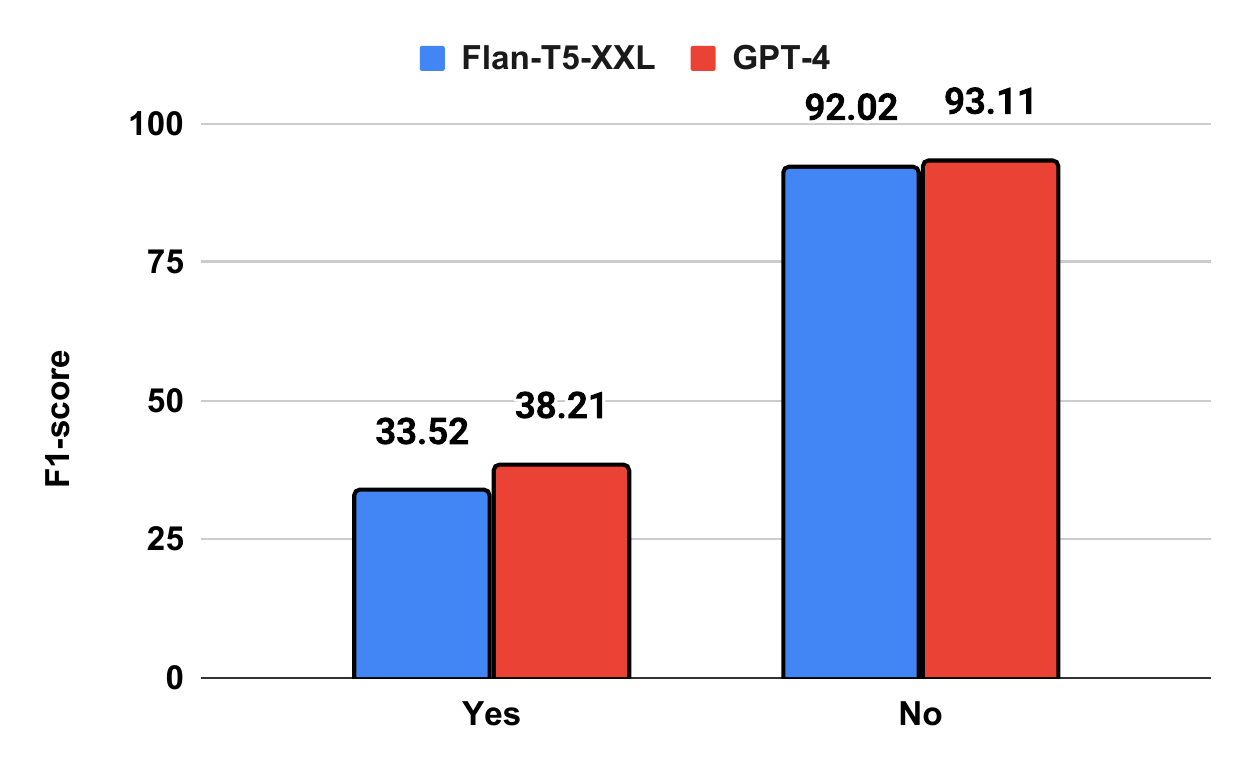}
  \caption{F1-score for social message detection task class wise}
  \label{class_wise_social_message}
\end{subfigure}%
\begin{subfigure}{.50\textwidth}
  \centering
  \includegraphics[width=\linewidth]{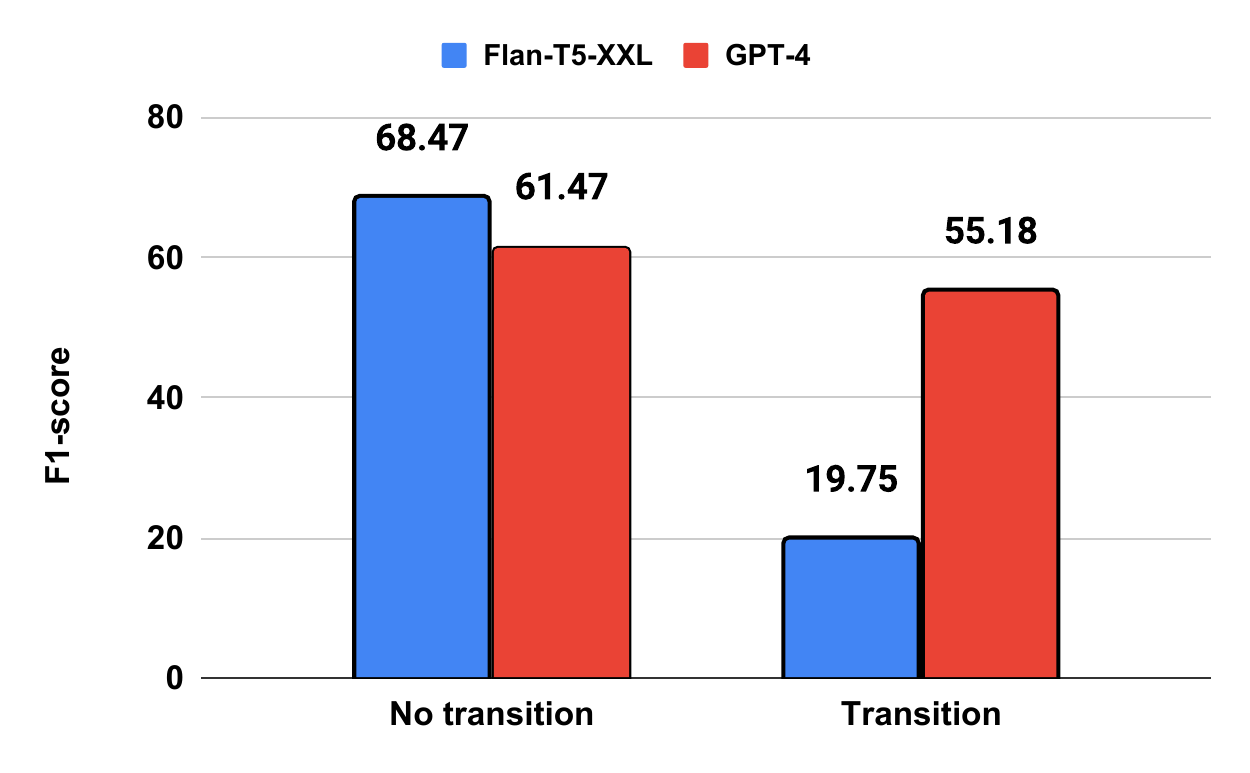}
  \caption{F1-score for tone transition detection task class wise}
  \label{class_wise_tone_transition}
\end{subfigure}
\caption{Comparisons between \texttt{Flan-T5-XXL} and \texttt{GPT-4} in terms of class-wise F1-score for the social message (Yes: Presence of social message, No: Absence of social message), Tone transition detection tasks.}
\label{tone_transition_social_message}
\end{figure*}
\subsubsection{Multimodal baselines}:
For our multimodal baselines, we use PerceiverIO \cite{Jaegle2021PerceiverIA} as the transformer encoder to fuse paired modalities (audio, text), (audio, visual), (text, visual). We use the publicly available implementation of PerceiverIO listed in: \textbf{\url{https://github.com/lucidrains/perceiver-pytorch}}. For the text modality, we use the pretrained \texttt{bert-base-uncased} model available with HuggingFace library \cite{Wolf2019HuggingFacesTS}.
We use the following hyperparameter choices for the audio-text perceiver $\mathbf{Tx_{AT}}$ model: \\\\
$\mathbf{Tx_{AT}}$: For $\mathbf{Tx_{AT}}$ with paired audio and text modalities, we use the following hyperparameters for tone transition and social message detection tasks:

\begin{itemize}
    \item \textbf{Batch size:} 16
    \item \textbf{Optimizer:} \texttt{Adam} \cite{kingma2014adam} (lr=1e-4)
    \item \textbf{Maximum audio sequence length}: 14
    \item \textbf{Maximum text sequence length}: 256
    \item \textbf{patience:} 5
    \item \textbf{queries\_dim:} 256
    \item \textbf{use\_queries:} False
    \item \textbf{depth:} 4
    \item \textbf{num\_latents:} 16
    \item \textbf{cross\_heads:} 1
    \item \textbf{latent\_heads:} 8
  \item \textbf{cross\_dim\_head:} 128
  \item \textbf{latent\_dim\_head:} 32
  \item \textbf{latent\_dim:} 256
  \item \textbf{weight\_tie\_layers:} False
  \item \textbf{seq\_dropout\_prob:} 0.1
\end{itemize}
For the topic categorization task, the hyperparameters remain the same for $\mathbf{Tx_{AT}}$ except the \textbf{cross\_dim\_head:}=32.
\\\\
$\mathbf{Tx_{AV}}$: For $\mathbf{Tx_{AV}}$ with paired audio and video modalities, we use the following hyperparameters for tone transition and social message detection tasks:

\begin{itemize}
    \item \textbf{Batch size:} 16
    \item \textbf{Optimizer:} \texttt{Adam} \cite{kingma2014adam} (lr=1e-5)
    \item \textbf{Maximum audio sequence length}: 14
    \item \textbf{Maximum video sequence length}: 35
    \item \textbf{patience:} 5
    \item \textbf{queries\_dim:} 256
    \item \textbf{use\_queries:} False
    \item \textbf{depth:} 4
    \item \textbf{num\_latents:} 16
    \item \textbf{cross\_heads:} 1
    \item \textbf{latent\_heads:} 8
  \item \textbf{cross\_dim\_head:} 32
  \item \textbf{latent\_dim\_head:} 32
  \item \textbf{latent\_dim:} 256
  \item \textbf{weight\_tie\_layers:} False
  \item \textbf{seq\_dropout\_prob:} 0.1
\end{itemize}
For the topic categorization task, the hyperparameters remain the same for $\mathbf{Tx_{AV}}$ except for the optimizer \texttt{Adam} \cite{kingma2014adam} (lr=1e-4).
\\\\
$\mathbf{Tx_{TV}}$: For $\mathbf{Tx_{TV}}$ with paired text and video modalities, we use the following hyperparameters for tone transition and social message detection tasks:

\begin{itemize}
    \item \textbf{Batch size:} 16
    \item \textbf{Optimizer:} \texttt{Adam} \cite{kingma2014adam} (lr=1e-4)
    \item \textbf{Maximum text sequence length}: 256
    \item \textbf{Maximum video sequence length}: 35
    \item \textbf{patience:} 5
    \item \textbf{queries\_dim:} 256
    \item \textbf{use\_queries:} False
    \item \textbf{depth:} 4
    \item \textbf{num\_latents:} 16
    \item \textbf{cross\_heads:} 1
    \item \textbf{latent\_heads:} 8
  \item \textbf{cross\_dim\_head:} 128
  \item \textbf{latent\_dim\_head:} 32
  \item \textbf{latent\_dim:} 256
  \item \textbf{weight\_tie\_layers:} False
  \item \textbf{seq\_dropout\_prob:} 0.1
\end{itemize}
For the topic categorization task, the hyperparameters remain the same for $\mathbf{Tx_{TV}}$ except for the optimizer \texttt{AdamW}  \cite{Loshchilov2017DecoupledWD} (lr=1e-4) and \textbf{Maximum text sequence length} =512.

\section{Results}

\subsection{Language-based reasoning}

\begin{figure*}[h!]
    \includegraphics[width=\textwidth]{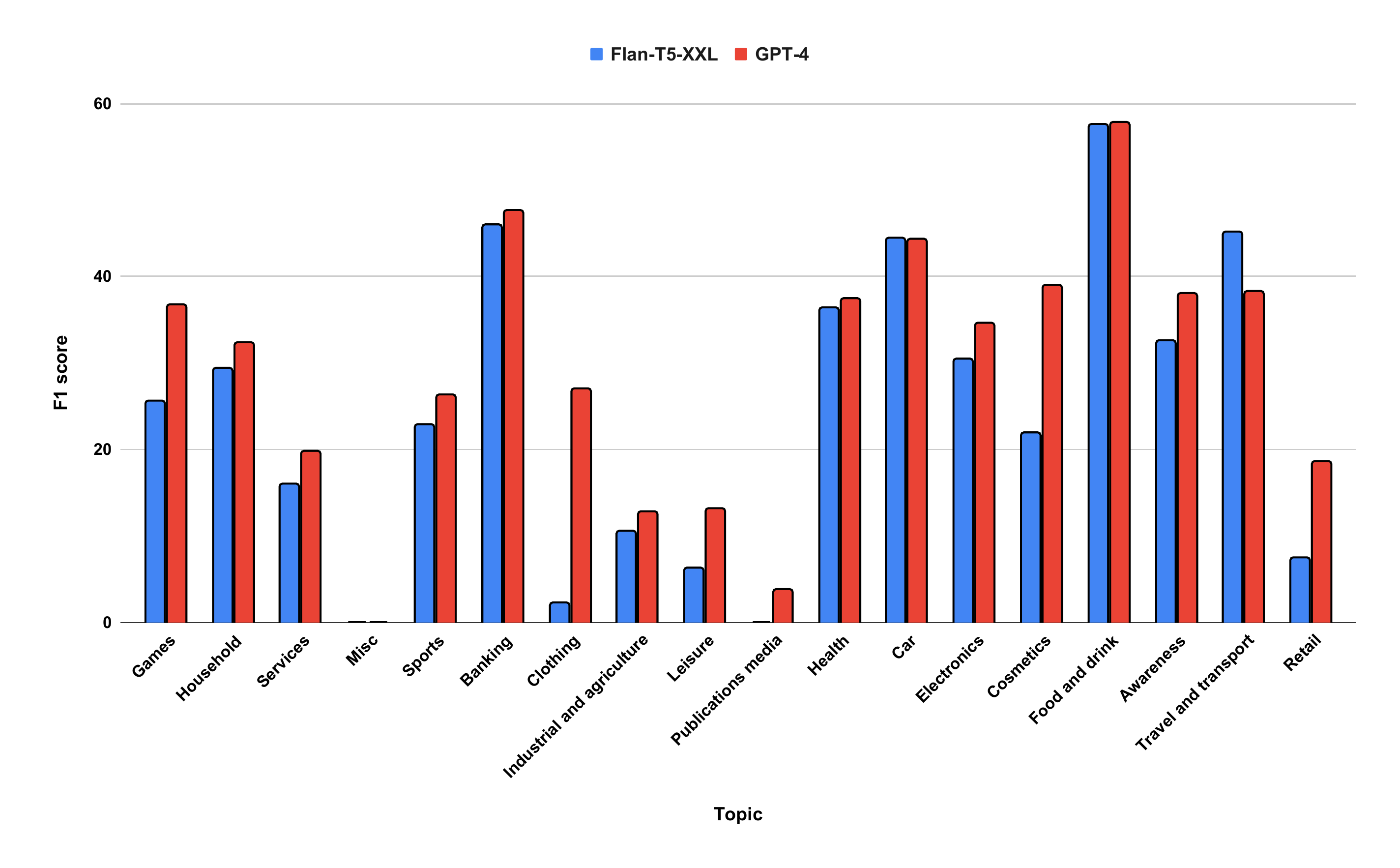}
    \caption{Comparisons between \texttt{Flan-T5-XXL} and \texttt{GPT-4} in terms of class-wise F1-score for the topic categorization task}
    \label{gpt_4_flan_t5_xxl_topic}
\end{figure*}

We provide class-wise comparisons between \texttt{Flan-T5-XXL} and \texttt{GPT-4} for the three benchmark tasks of social message, tone transition detection, and topic categorization. From Fig \ref{tone_transition_social_message} (a), we can see that \texttt{GPT-4} obtains a higher F1-score (\textbf{38.21\%}) as compared to \texttt{Flan-T5-XXL} (\textbf{33.52\%}) for the \textbf{Yes} class signifying the presence of social message in a complete zero-shot setting. Further for the tone-transition task, \texttt{GPT-4} obtains a significantly higher F1-score (\textbf{55.18\%}) for the \textbf{Transition} class than \texttt{Flan-T5-XXL} (\textbf{19.75\%}). In terms of topic categorization, we can see from Fig \ref{gpt_4_flan_t5_xxl_topic} that \texttt{GPT-4} performs better than \texttt{Flan-T5-XXL} in the case of all categories, except for the \textbf{Awareness} class.  For the minority topic categories i.e. \textbf{Retail}, \textbf{Publications media}, \textbf{Industrial and Agriculture}, \texttt{GPT-4} performs slightly better than \texttt{Flan-T5-XXL}. The poor performance of \texttt{GPT-4} and \texttt{Flan-T5-XXL} in the \textbf{Misc} category can be attributed to the grouping of multiple diverse subcategories like \textbf{Petfood}, \textbf{Business and equipment}, \textbf{Politics} into a single large category. Future work will involve the usage of expanded topic taxonomy to mark the transcripts with respective categories by incorporating reasoning mechanisms like chain of thought prompting \cite{Wei2022ChainOT}.

\subsection{Unimodal and multimodal baselines}
\begin{figure*}[h!]
\begin{subfigure}{.50\textwidth}
  \centering
  \includegraphics[width=\linewidth]{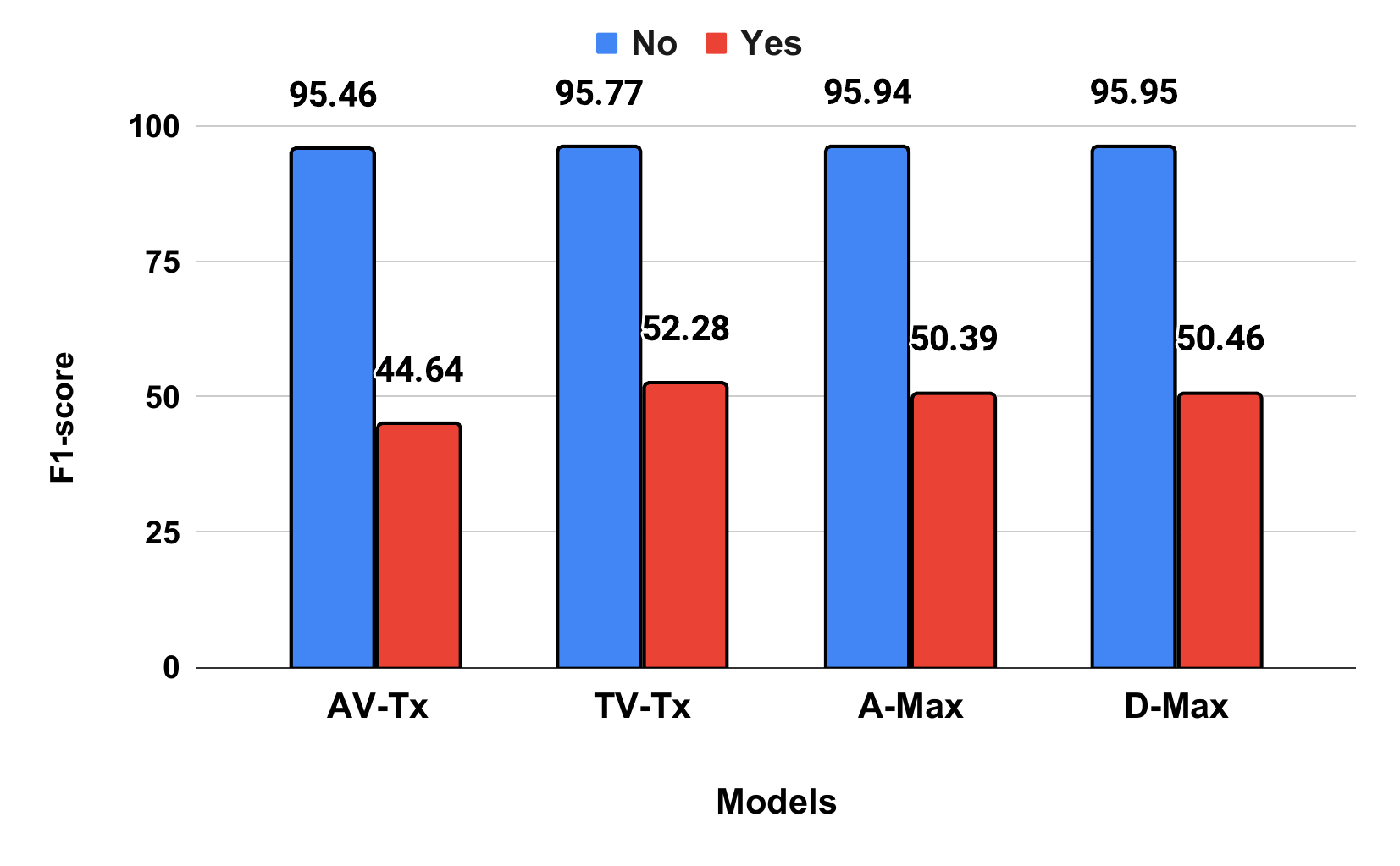}
  \caption{F1-score for social message detection task class-wise}
  \label{class_wise_social_message_multimodal}
\end{subfigure}%
\begin{subfigure}{.50\textwidth}
  \centering
  \includegraphics[width=\linewidth]{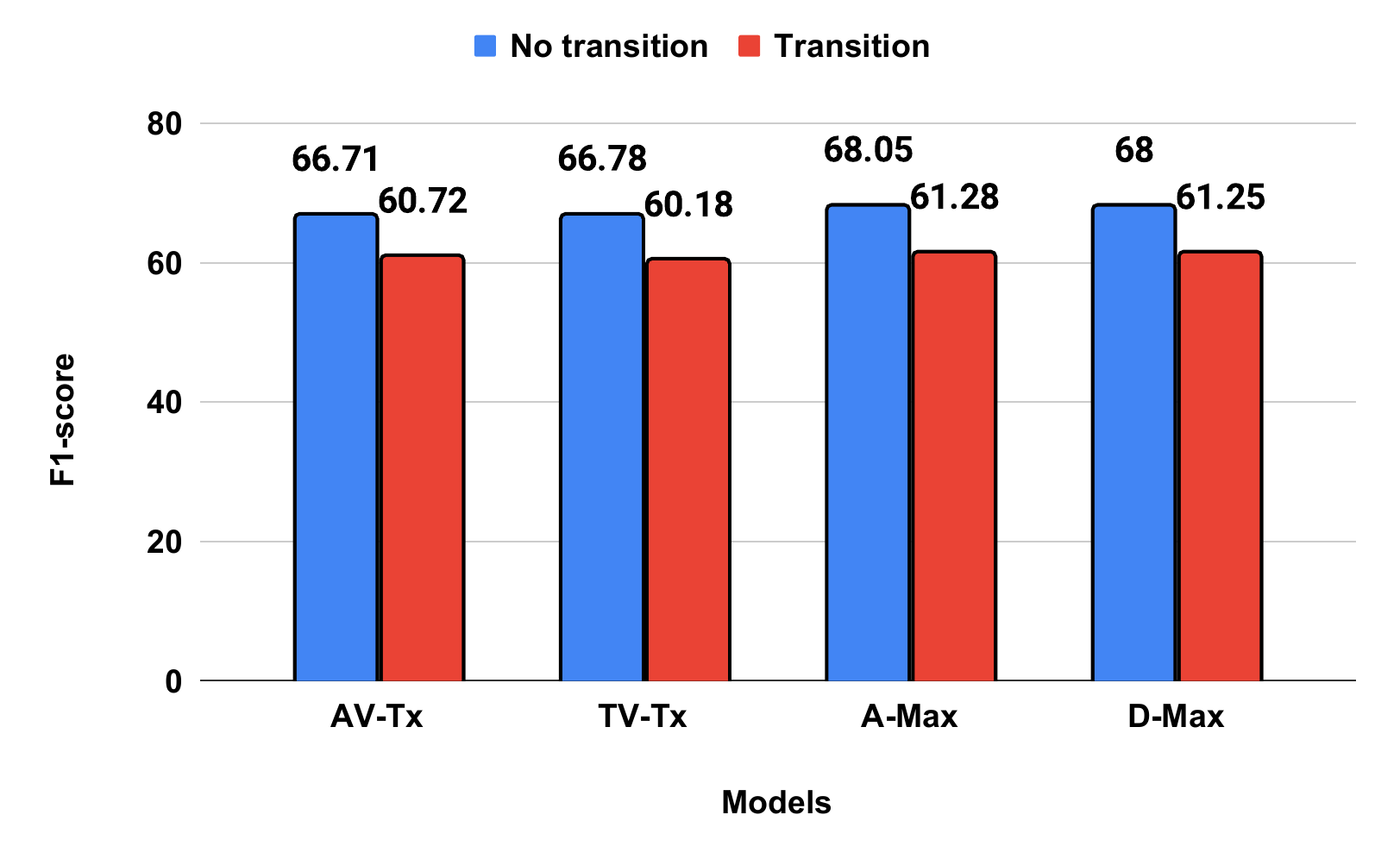}
  \caption{F1-score for tone transition detection task class-wise}
  \label{class_wise_tone_transition_multimodal}
\end{subfigure}
\caption{Comparisons between $\mathbf{Tx_{AV}}$(\texttt{AV-Tx}), $\mathbf{Tx_{TV}}$(\texttt{TV-Tx}), \texttt{A-Max}, \texttt{D-Max} in terms of class-wise F1-score for the social message (Yes: Presence of social message, No: Absence of social message), Tone transition detection tasks. Average of 5 runs considered for the F1-score.}
\label{tone_transition_social_message_multimodal}
\end{figure*}

\begin{figure*}[h!]
    \includegraphics[width=\textwidth]{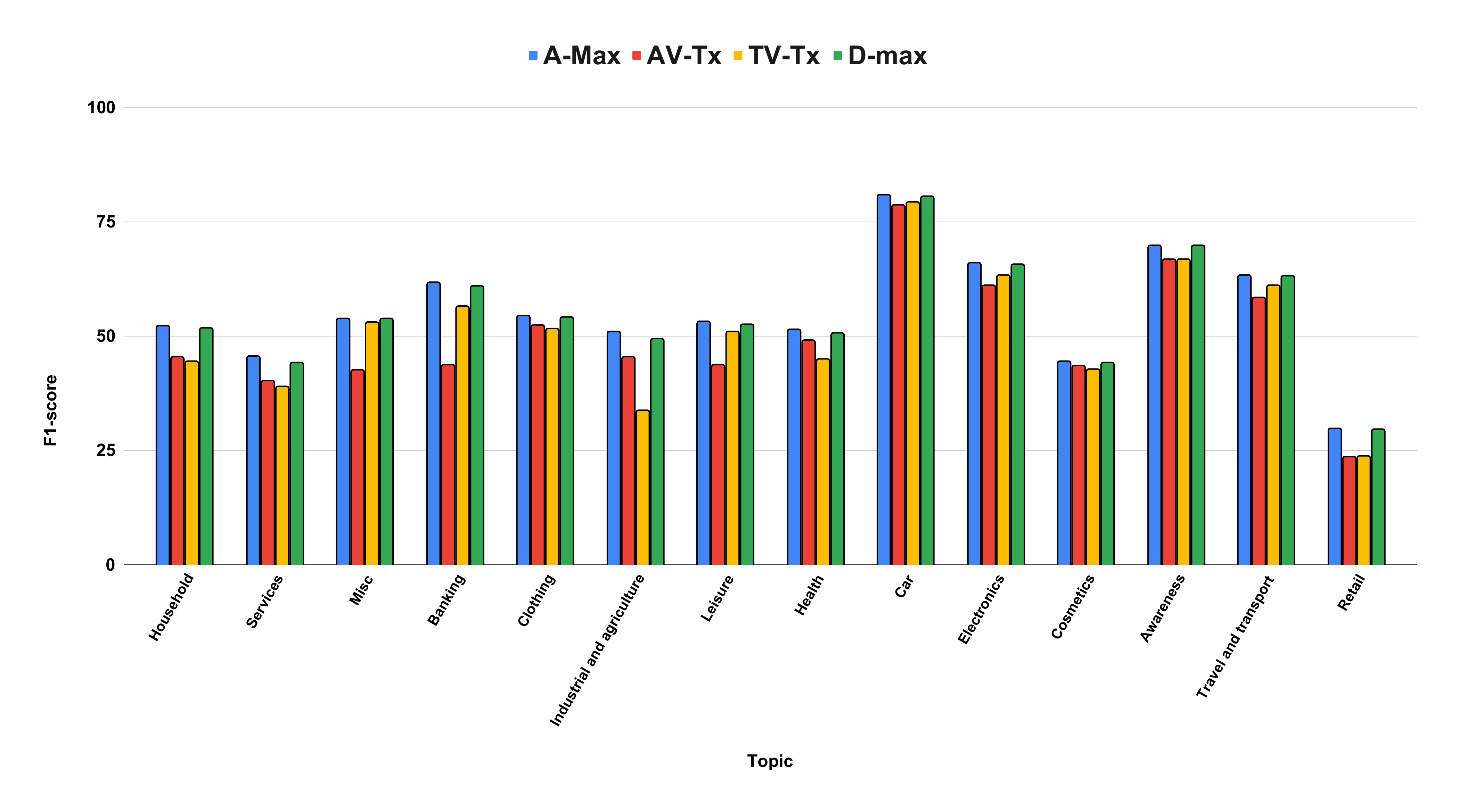}
    \caption{Comparisons between $\mathbf{Tx_{AV}}$ (\texttt{AV-Tx}), $\mathbf{Tx_{TV}}$(\texttt{TV-Tx}), \texttt{A-Max}, \texttt{D-Max} in terms of class-wise F1-score for the topic categorization task. Average of 5 runs is considered for the F1-score.}
    \label{topic_multimodal}
\end{figure*}

From Fig \ref{tone_transition_social_message_multimodal} (b), we can see that average-Max (\textbf{A-Max}) and dual-Max (\textbf{D-Max}) logit fusion of $\mathbf{Tx_{AV}}$ and  $\mathbf{Tx_{TV}}$ improves the average F1-score to \textbf{61.28\%} and \textbf{61.25\%} respectively for the \textbf{Transition} class. However, we observe from Fig \ref{tone_transition_social_message_multimodal} (a), that both average-Max (\textbf{A-Max}) and dual-Max (\textbf{D-Max}) fusion strategies do not help for social message detection (i.e. \textbf{Yes} class). For topic categorization, as seen from Fig \ref{topic_multimodal}, we obtain consistent improvements across 14 categories (out of 18) for both average-Max (\textbf{A-Max}) and dual-Max (\textbf{D-Max}) fusion schemes. Both  (\textbf{A-Max}) and dual-Max (\textbf{D-Max}) fusion schemes perform similarly in terms of average F1-score except for the \textbf{Food and Drink} class. For \textbf{Food and Drink}, \textbf{D-Max} fusion results in a worse performance as compared to $\mathbf{Tx_{AV}}$ and $\mathbf{Tx_{TV}}$.




\end{document}